\pdfoutput=1
\documentclass{article} % For LaTeX2e
\usepackage[final]{colm2026_conference}

\usepackage{microtype}
\usepackage{url}
\usepackage{booktabs}
\usepackage{graphicx}
\usepackage{amsmath,amssymb,amsfonts,bm}
\usepackage{algorithm}
\usepackage{algorithmic}
\usepackage{newfloat}
\DeclareFloatingEnvironment[name=Prompt]{floatquote}
\usepackage{listings}
\usepackage{enumitem}
\usepackage{wrapfig}
\usepackage{array}
\usepackage{makecell}
\usepackage{multirow}
\usepackage{lineno}
\usepackage{xcolor}
\usepackage{hyperref}

\definecolor{darkblue}{rgb}{0, 0, 0.5}
\hypersetup{colorlinks=true, citecolor=darkblue, linkcolor=darkblue, urlcolor=darkblue}

\title{PACIFIC: Can LLMs Discern the Psychometric Traits Influencing Your
Preferences? Personality-Driven Preference Alignment in LLMs}

\author{Tianyu Zhao$^{*}$, 
Siqi Li \thanks{Equal contribution. Correspondence to Tianyu Zhao.}, 
Yasser Shoukry \& 
Salma Elmalaki \\
Department of Electrical Engineering and Computer Science \\
University of California, Irvine\\
Irvine, CA 92697, USA \\
\texttt{\{tzhao15, siqi.li, yshoukry, selmalak\}@uci.edu} \\
}

\begin{document}

\ifcolmsubmission
\linenumbers
\fi

\maketitle

\begin{abstract}
User preferences are increasingly used to personalize Large Language Model (LLM) responses, yet reliably leveraging preference signals remains under-explored. In practice, preferences can be noisy, incomplete, or even misleading, which can degrade answer quality when applied naively. Motivated by the observation that stable personality traits shape everyday preferences, we introduce \textbf{PACIFIC} (\textbf{P}reference \textbf{A}lignment for \textbf{C}hoices \textbf{I}nference via \textbf{F}ive-factor \textbf{I}dentity \textbf{C}haracterization), a personality-driven preference alignment framework that uses Big-Five (OCEAN) traits as a principled ''latent'' signal for organizing and reasoning over user preference history. To systematically evaluate this framework, we construct a psychometrics-based dataset containing $1{,}200$ preference--query pairs spanning diverse domains (e.g., travel, movies, and education), with comprehensive coverage of high and low Big-Five trait directions. Extensive experiments show that trait-aligned preferences substantially improve personalized QA: given clean, trait-aligned context, LLMs reach near-ceiling accuracy (up to $99\%$), confirming that reasoning capability is not the bottleneck. The challenge is that real histories are mixed-trait and unlabeled. We show the true bottleneck is retrieval: a persona-aware contrastive retriever (PiRAG) raises label-free accuracy from $30\%$ to $43\%$ over standard semantic retrieval, without any trait annotations at inference. Dataset:{ \raisebox{-0.3em}{\includegraphics[height=1.1em]{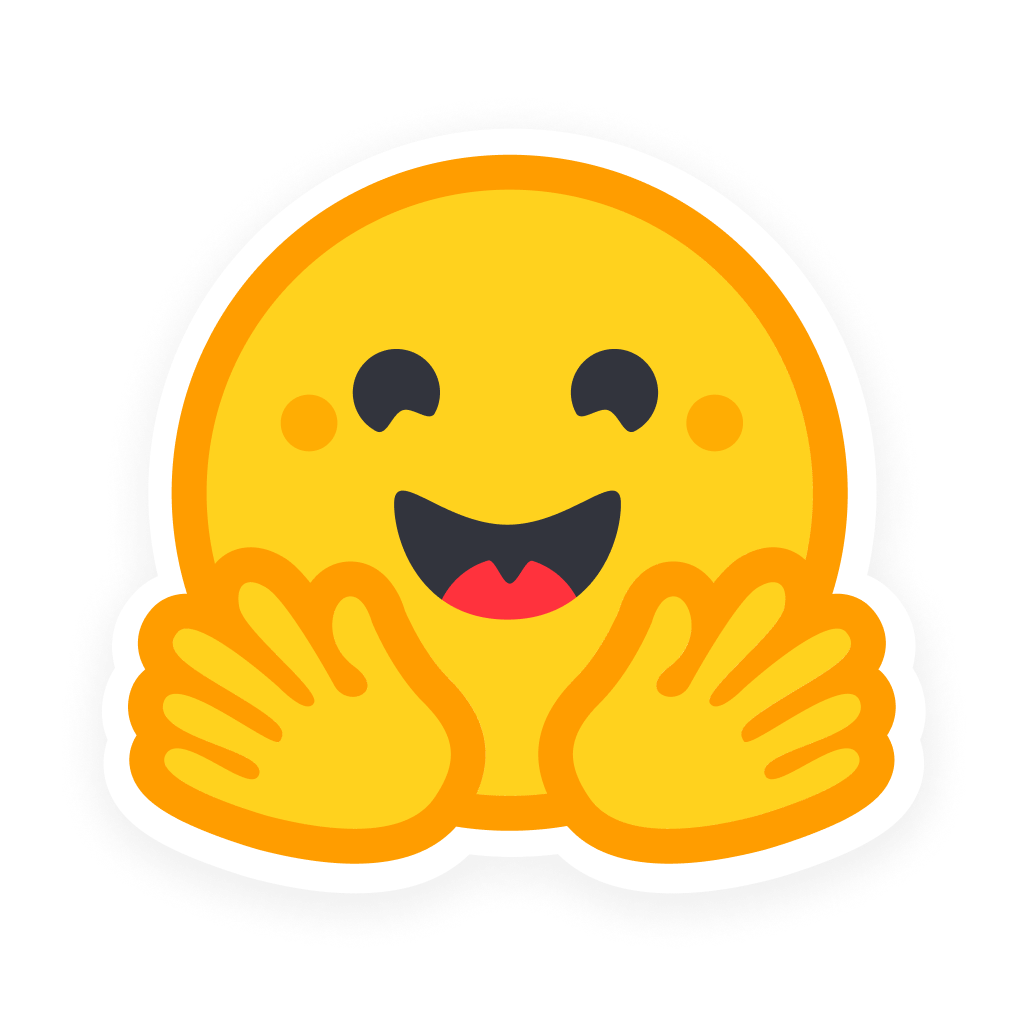}}} \url{https://huggingface.co/datasets/TylerZ0931/PACIFIC-big-five-trait-preferences}

\end{abstract}

% \begin{figure}[!t]
%     \centering
%       \includegraphics[width=\linewidth]{figs/motivation.pdf}
%         \caption{Motivation: Topic matching may fail to capture a user’s underlying preferences, whereas personality-based reasoning enables better preference inference in a new query. The illustration above shows a simplified example. }
%         % - \salma{I tried this flow of preference in gemini with different variations and it gave me the right answer that is personality aligned. I even tested it with different preference set using a trap query (Keyword \& Concept Hijack) and again it didn't fail. We need something that is more implicit to put as an example.}
%     \label{fig:motivation}
%     \vspace{-0.2in}
% \end{figure}

\begin{wrapfigure}{R}{0.52\linewidth}
    \centering
    \vspace{-0.5em}
    \includegraphics[width=\linewidth]{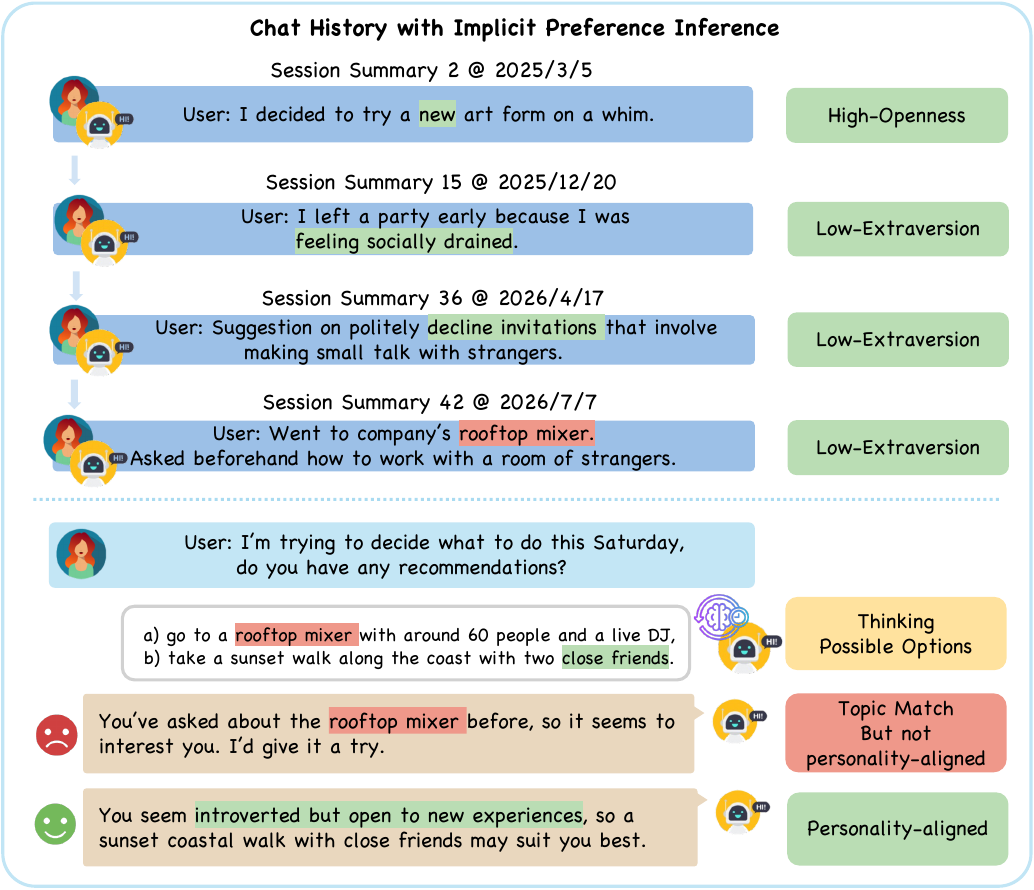}
    \vspace{-0.3in}
    \caption{Motivation: Topic matching may fail to capture a user’s underlying preferences, whereas personality-based reasoning enables better preference inference in a new query. The illustration above shows a simplified example. }
    \label{fig:motivation}
    \vspace{-0.3in}
\end{wrapfigure}

\section{Introduction}
%As LLMs become widely used, people increasingly rely on them for recommendations in everyday settings such as education, travel, and entertainment. This growing dependence has fueled interest in chatbots that can remember a user's likes and constraints---for example, suggesting restaurants that match dietary needs or recommending shows aligned with personal tastes. While modern systems such as Gemini~\citep{team2023gemini} and GPT-4~\citep{achiam2023gpt} have substantially improved language understanding and generation, delivering reliable personalization at scale and over long conversations remains challenging. Prior work in the literature~\citep{zhao2025llms} finds that preference-following accuracy can fall below $10\%$ after only $10$ turns (about $3k$ tokens) for many models, and performance continues to decline as the conversation grows, even with stronger prompting or retrieval. This gap arises for several reasons. Preferences mentioned earlier are often not applied later, especially when the context becomes long. User preferences can also be incomplete, inconsistent, or noisy, which can lead to poor recommendations. Moreover, models may incorrectly infer preferences that the user never stated. Together, these issues make it brittle to depend on a model to track and use a large set of detailed preferences over time.

As Large Language Models (LLMs) become increasingly integrated into everyday applications, users naturally expect them to infer their preferences when facing new questions, even when the relevant preference has never been explicitly discussed. At the same time, continued interaction produces rapidly growing user history, making long-term personalization more difficult. Although modern systems such as Gemini~\citep{team2023gemini} and GPT-4~\citep{achiam2023gpt} have substantially improved language understanding and generation, they still struggle to identify and use the right preference signals over long conversations. Prior work~\citep{zhao2025llms} finds that preference-following accuracy can fall below $10\%$ after only $10$ turns (about $3k$ tokens) for many models, and performance continues to decline as the conversation grows.

Interestingly, humans often process long interaction history in the opposite way: as interactions accumulate, their understanding of another person can become more reliable, abstract, and generalizable. Consider the example in Fig.~\ref{fig:motivation}: a friend you know tends to \emph{try new experiences, feel socially drained at parties, and avoid small talk with strangers}. If they later ask for suggestions for a Saturday activity, a topic-based strategy may focus on their earlier mention of a \emph{rooftop mixer} and recommend attending it. As a human, however, you may instead easily infer from the interactions that they would likely prefer trying something new alone or with a few close friends. These seemingly unrelated past behaviors reveal a consistent tendency toward introversion. In other words, people do not recall previous conversation details; instead, they form a higher-level understanding of another person's personality and use it to generalize to new situations.

%We therefore take a complementary approach. \textbf{\emph{Rather than requiring an LLM to remember many specific preferences, we use personality traits as a more stable signal that helps organize and interpret preferences across domains.}} Because personality tends to be consistent, it can support more reliable decisions even when individual preferences are sparse, outdated, or noisy. In this way, personality-guided personalization can be both more efficient and more robust than attempting to store and retrieve every preference verbatim.

This suggests that effective LLM personalization may require more than storing and retrieving preference statement histories. Instead, LLMs should learn to use these histories to construct a more structured representation of the user and use that representation to infer user preferences and give recommendations for questions from new domains.

We explore \emph{Big-Five personality} as such a representation. The Big-Five personality traits -- Openness, Conscientiousness, Extraversion, Agreeableness, and Neuroticism (OCEAN) -- capture relatively stable behavioral tendencies and are associated with systematic patterns in preferences and decision making~\cite{li-etal-2025-big5}. We propose that, \textbf{\emph{rather than requiring an LLM to remember many specific preferences, we use Big-Five traits as a more stable signal that helps organize and interpret preferences across domains.}} This allows preferences to be connected through shared psychometric alignment rather than semantic similarity alone. We therefore ask: \emph{Can personality serve as a useful latent representation for organizing user preferences and improving preference following in LLMs?}

We first test our hypothesis on \textsc{PrefEval} \cite{zhao2025llms}, an independently constructed preference-following benchmark of 1,000 synthesized user preferences, questions, and ground-truth answers. Our results (Tab.\ref{table:experiment_prefeval}) show that personality-aligned preference information improves preference following of LLMs.
% \salma {refer to where are those results in the paper FIXED} 

\begin{figure*}[h]
  %\vskip -0.2in
  \begin{center}
    \centerline{\includegraphics[width=1.0\textwidth]{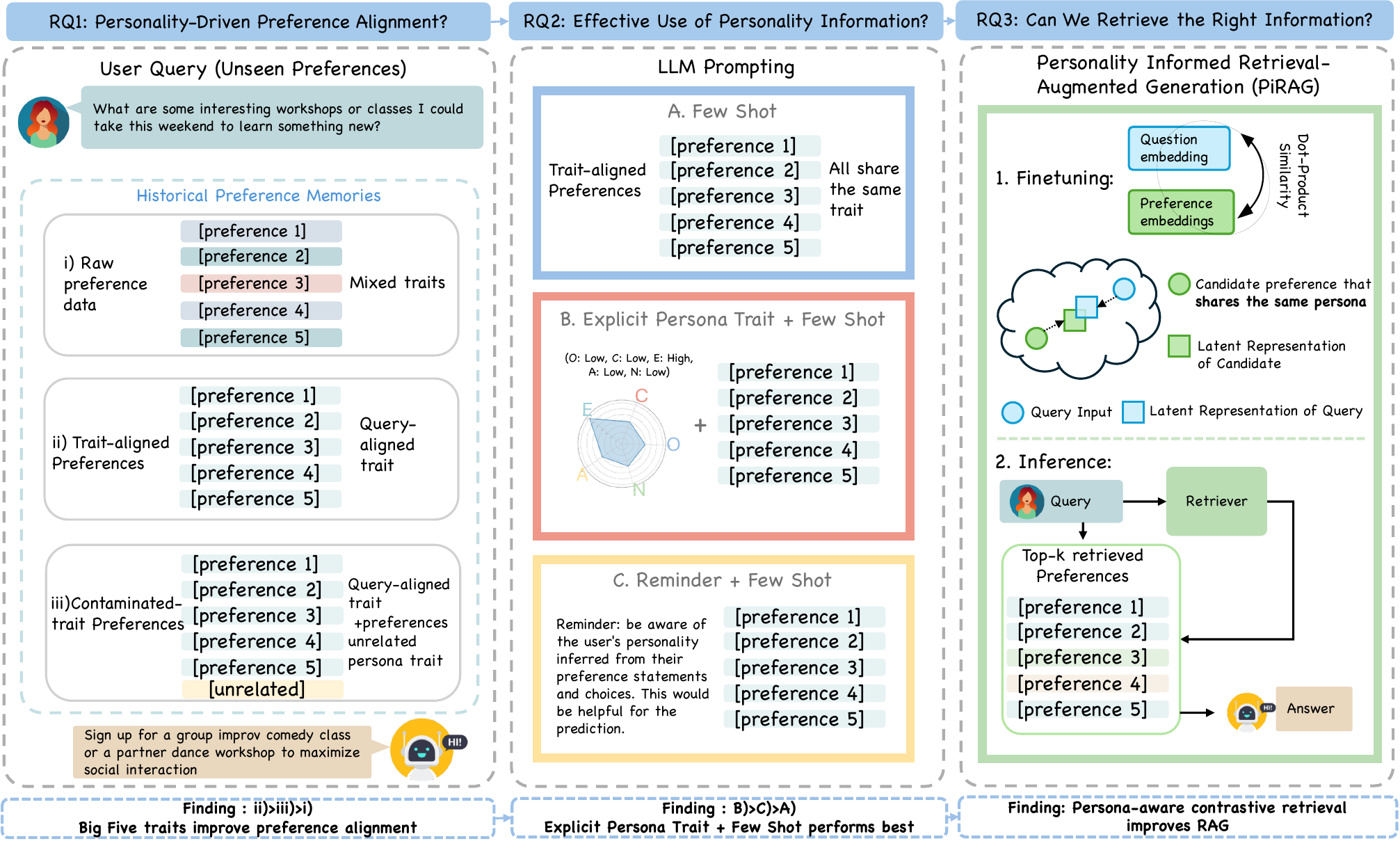}}
    \vspace{-0.1in}
    \caption{
\textbf{Overview of our PACIFIC framework.}
\textbf{RQ1 (left)} evaluates whether personality provides a useful latent structure for preference alignment by comparing mixed-trait, trait-aligned, and contaminated trait-aligned preference histories for predicting an unseen user preference.
\textbf{RQ2 (middle)} investigates how personality information should be incorporated into the LLM prompting, comparing few-shot alone, with explicit persona traits, and with implicit persona reminders.
\textbf{RQ3 (right)} studies whether personality-aligned information can be automatically retrieved using Personality-informed RAG (PiRAG).
}
 % \salma{This figure was referenced after figure 3 in the text. It has to be referenced before figure 3,FIXED}
\label{fig:method-overview}

  \end{center}
\vspace{-0.3in}
\end{figure*}

However, \textsc{PrefEval} was not designed for systematic personality analysis, and its preferences are unevenly distributed across the ten high/low Big-Five trait directions. To enable controlled comparisons, we construct a balanced psychometrics-based dataset and investigate the following research questions illustrated in Fig. \ref{fig:method-overview}: 
%\salma{refer to figure 2 here somehow FIXED}:

\noindent\hangindent=10pt\hangafter=1
\textbullet\ \textbf{RQ1 (Personality-Driven Preference Alignment).}
Can Big Five traits provide a useful user abstraction that enables LLMs to infer preferences in a new query especially when that preference has never been explicitly stated?\par
\noindent\hangindent=10pt\hangafter=1
\textbullet\ \textbf{RQ2 (Effective Use of Personality Information).}
How should personality information be incorporated to maximize its benefit for preference prediction?\par
\noindent\hangindent=10pt\hangafter=1
\textbullet\ \textbf{RQ3 (Retrieving Personality-Aligned Information).}
Without explicit personality labels, can a retriever identify personality-aligned memories that support the prediction of unseen preferences?\par

% \begin{itemize}[leftmargin=*, itemsep=0pt, topsep=0pt]
%     \item \textbf{RQ1 (Personality-Driven Preference Alignment):}
%     Can Big Five traits provide a useful user abstraction that enables LLMs to infer preferences in a new query especially when that preference has never been explicitly stated?

%     \item \textbf{RQ2 (Effective Use of Personality Information):}
%     How should personality information be incorporated to maximize its benefit for preference prediction?

%     \item \textbf{RQ3 (Retrieving Personality-Aligned Information):}
%      Without explicit personality labels, can a retriever identify personality-aligned memories that support the prediction of unseen preferences?
% \end{itemize}

% \begin{itemize}[leftmargin=*, itemsep=1pt, topsep=1pt]
%     \item \textbf{RQ1 (Understanding or Memory Matching):} Given a chat history with some implicit preference inference, 
%     Can LLMs infer a user's preference in a new query,  when that preference has never been explicitly stated?

%     %Can LLMs infer a user's preference in a new situation from indirect evidence, even when that preference has never been explicitly stated? 

%     \item \textbf{RQ2 (Personality as a User Model):}
%     Can Big Five traits provide a useful abstraction of the user that organizes past preferences and improves prediction of unseen choices?

%     \item \textbf{RQ3 (Retrieving the Right Memories):}
%     Without explicit personality labels, can a retriever identify personality-aligned memories that support the prediction of unseen preferences?
    
% \end{itemize}
We make three contributions: (1) We show that personality traits improve preference following in personalized question answering. (2) We propose the \textbf{PACIFIC} framework, which incorporates personality-aligned preferences during generation. LLMs reach near-ceiling accuracy given clean trait-aligned context, and since real histories are messy and unlabeled, our persona-aware retriever (PiRAG) improves label-free accuracy from $30\%$ to $43\%$ without trait annotations. (3) We introduce a psychometrics-based dataset with $1{,}200$ preference statements and conversations spanning diverse domains (e.g., travel, movies, education), annotated with Big-Five (OCEAN) trait directions \citep{briggs1992assessing,de2000big,goldberg2013alternative}. 
% We propose the \textbf{PACIFIC} framework, which incorporates personality-aligned preferences during generation and improves accuracy from $29.25\%$ to $76\%$, suggesting that personality-guided preference retrieval benefits preference alignment.
% \salma{Be consistent. Later you call it Psychometrics-based dataset not personality-label preference dataset. FIXED }

%The paper is organized as follows: \autoref{sec:pacificdataset} introduces \textsc{PACIFIC} dataset. \autoref{sec:framework} proposes our persona-driven preference-following framework. \autoref{sec:evaluation} details the experiment setup and evaluation results. \autoref{sec:related} highlights the related work. The paper concludes with discussion and conclusion in \autoref{sec:discuss}.

% \vspace{-8mm}

\section{Persona Trait Boosts Preference Following}
\vspace{-2mm}
We evaluate our hypothesis using \textsc{PrefEval}~\citep{zhao2025llms}, an existing benchmark for preference following. As illustrated by the blue steps in Fig.~\ref{fig:pacific-dataset-overview}, \textsc{PrefEval} covers 20 topics (listed in Suppl.~\ref{sec:topics}) and consists of synthesized user preferences $p_i$, downstream multiple-choice questions $q_i$, and four candidate answers $\{c_{i,k}\}_{k=1}^{4}$. For each question, $a_i \in [1,4]$ denotes the index of the preference-adhering ground-truth answer, i.e., $c_{i,a_i}$. An LLM-based judge is then used to filter out invalid or insufficiently challenging instances, resulting in a final set of 1,000 preference--question--answer triples $(p_i,q_i,a_i)$.

\subsection{Annotate Trait Label}
\label{subsec:Personality Trait Annotation}
\begin{figure*}[!t]
% \vspace{-5mm}
  \vskip -0.2in
  \begin{center}
    \centerline{\includegraphics[width=1.0\textwidth]{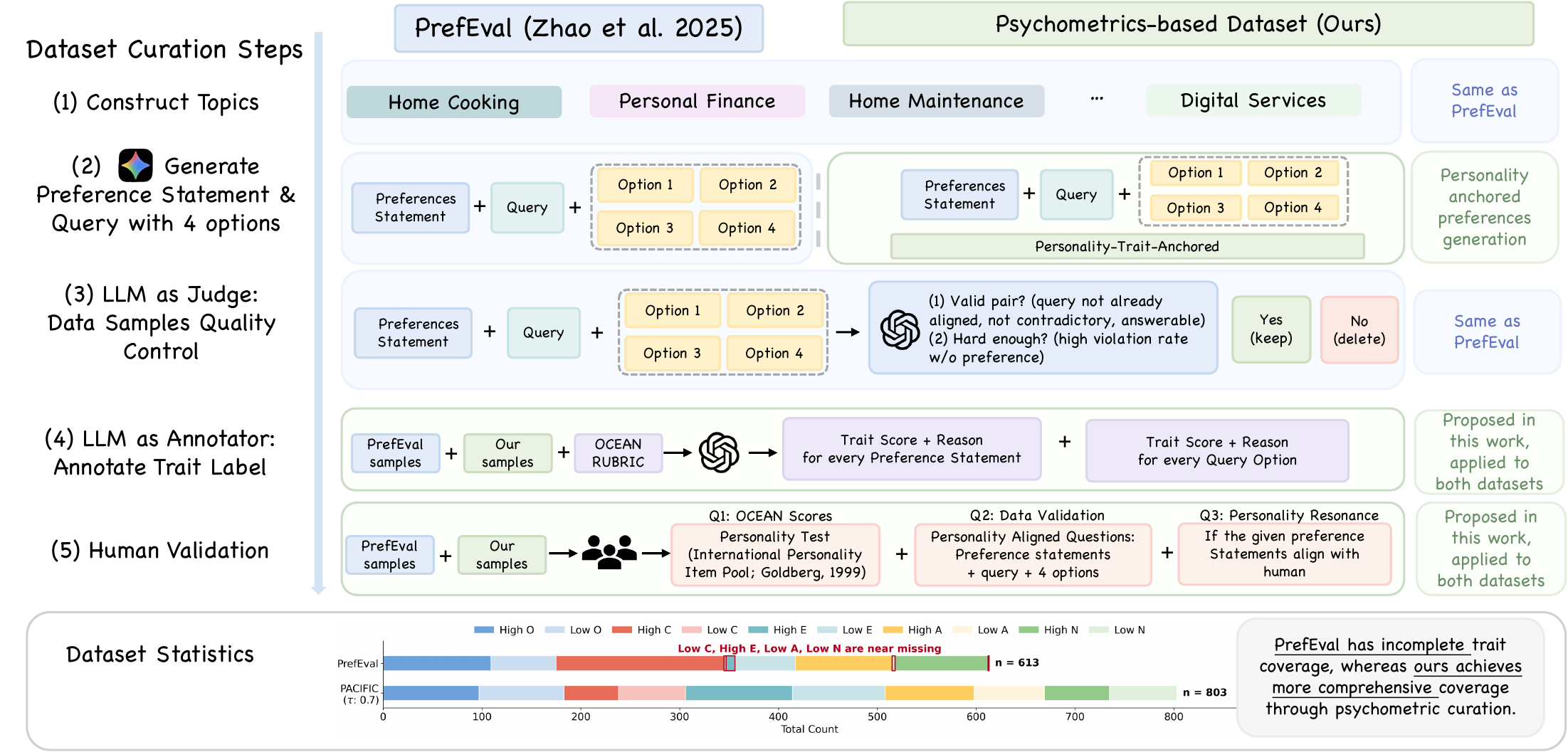}}
    \vspace{-0.1in}
    \caption{
    \textbf{Our psychometrics-based dataset curation pipeline.} We compare the \textsc{PrefEval} \cite{zhao2025llms} pipeline with our psychometrics-based extension. Our five-step construction process begins by (1) retaining the 20-topic foundation from \textsc{PrefEval}. (2) To ensure comprehensive trait coverage for targeted analysis, we generate personality-anchored preference statements and queries rather than deriving them directly from topics. (3) Consistent with \textsc{PrefEval}, we filter out statements that language models fail to predict accurately to maintain data quality. (4) We then annotate all statements and query options with OCEAN trait labels and corresponding reasons, and (5) validate them through human assessments of personality, preference alignment, and personality resonance. As illustrated in the bar plot, the resulting dataset achieves significantly more comprehensive coverage across both high- and low-level Big Five traits compared to the \textsc{PrefEval} baseline.
    }
    \label{fig:pacific-dataset-overview}
  \end{center}
\vspace{-0.3in}
\end{figure*}

Importantly, we leave these original preferences, questions, and answer labels unchanged and augment each preference statement and each question choice with Big-Five personality annotations, as shown in Step 4 of Fig.~\ref{fig:pacific-dataset-overview}. Specifically, each preference $p_i$ and answer choice $c_{i,k}$ is scored on a 7-point Likert scale along the five personality dimensions $\{O,C,E,A,N\}$, where 1 indicates a very low expression of a trait and 7 indicates a very high expression. 

\subsubsection{Dual-Strategy Annotation}
\label{subsec:Dual-Strategy Annotation}
To ensure psychometric validity, we employ a Dual-Strategy Annotation Protocol separating identity-expressive traits $\{O,C,E,A\}$ from need-compensatory traits $\{N\}$. Let $x$ denote an annotated context, i.e., either a preference $p_i$ or an answer choice $c_{i,k}$. Each such $x$ is annotated with  (i) a trait score vector $\mathbf{s}(x)\in[1,7]^5$ over $\{O,C,E,A,N\}$ and (ii) a confidence vector $\boldsymbol{\gamma}(x)\in[0,1]^5$ indicating the trait score reliability.
%\salma{What is $x$ here? is it the choice or the preference? Is it the text or the query? or prompt-response pair? or what is it? . Either use the traits O, C, E, A, N in math (\$ \$)environment or don't, but not mix. FIXED}

\paragraph{The Mirroring Strategy (O, C, E, A)}

For Openness, Conscientiousness, Extraversion, and Agreeableness, we adopt a Mirroring Strategy from Self-Congruity Theory ~\citep{sirgy1985using, sirgy2018self}, which posits that individuals prefer products, activities, or content that reinforce their established self-concept. Users with these traits predominantly seek \textbf{\textit{Alignment}}; for example, a highly Conscientious user prefers highly structured tools, while a highly Open user prefers novel and complex experiences. Scoring Criteria for O, C, E, A are in Supp.~\ref{app:criteria}.

%  \vspace{-2mm}
% \subparagraph{Scoring Criteria for O, C, E, A:} To operationalize this, we define the annotation scale based on the \textit{spectrum} of the trait characteristics present in the preference statement:
% \begin{itemize}[nosep, leftmargin=*]
%     \item \textit{Score 1-3 (Low Trait Expression)}: The preference reflects the characteristics associated with the low end of the trait spectrum (e.g., Routine/Tradition for Openness). 
%     \item \textit{Score 4 (Neutral)}: The preference shows no strong directional alignment with the trait.
%     \item \textit{Score 5-7 (High Trait Expression): } The preference reflects the characteristics associated with the high end of the trait spectrum (e.g., Novelty/Complexity for Openness).
% \end{itemize}

\paragraph{The Compensatory Strategy (N)}

For Neuroticism (N), the Mirroring Strategy is psychometrically flawed. A user with High Neuroticism (anxiety-prone) does not seek ``High Neuroticism'' experiences (risk/stress); rather, they seek \textbf{\textit{Safety}} \citep{tamir2005don, hirsh2012personalized}. Therefore, we apply a Compensatory Strategy based on the Safety vs. Efficiency trade-off.
High-N users view preferences as a defensive mechanism, prioritizing \textit{Psychological Cushioning} (guarantees, predictability) to mitigate anxiety. Conversely, Low-N users possess high emotional resilience and capacity for stress. They are outcome-oriented and prioritize efficiency over comfort, often perceiving unrequested safety buffers as \textit{coddling}. For example, consider a traveler choosing between a 45-minute and a 3-hour layover. A Low-N user would prefer the tight 45-minute layover to save time, accepting the risk of rushing as a calculated trade-off, whereas a High-N user would prioritize the 3-hour layover for the peace of mind it provides. Scoring criteria for N are provided in Suppl.~\ref{app:criteria}.

%  \vspace{-2mm}
% \subparagraph{Scoring Criteria for N:} To operationalize this, we defined the annotation scale for Neuroticism based on the provision of \textit{Safety vs. Efficiency}:
% \begin{itemize}[nosep, leftmargin=*]
%     \item \textit{Score 1-3 (High Efficiency / High Risk):} The preference prioritizes performance, efficiency, brutal truths, and high stakes. (Target: Low-N Users). 
%     \item \textit{Score 4 (Neutral)}: A balance between safety and performance.
%     \item \textit{Score 5-7 (High Safety / High Comfort):} The preference prioritizes guarantees, emotional regulation, and risk avoidance. (Target: High-N Users).
% \end{itemize}

\paragraph{Trait Label Mapping}

To simplify downstream analysis and reporting, we discretize continuous scores by mapping each context $x$ (e.g., a preference $p_i$ or a choice $c_{i,k}$) to a trait label. Let $\mathcal{D} = \{T^{\text{H}}, T^{\text{L}} \mid T \in \{O, C, E, A, N\}\}$ denote the set of high and low Big Five trait indicators. For each trait, the continuous score $s_t(x) \in [1, 7]$ is mapped to a discrete intensity $d_t(x)$, which is $\text{low}$ if $s_t(x) < 4$, $\text{high}$ if $s_t(x) > 4$, and $\text{unclear}$ if $s_t(x) = 4$. We then assign a categorical label $t(x) \in \mathcal{D} \cup \{\text{unclear}\}$ according to $d_t(x)$: a distinctly high or low intensity yields the corresponding indicator $T^{\text{H}}$ or $T^{\text{L}}$, and otherwise $t(x) = \text{unclear}$.

%  \paragraph{Trait Label Mapping}
% To simplify downstream analysis and reporting, we discretize continuous scores by mapping each text x to a single categorical trait label:
% \[
% d_t(x)=
% \begin{cases}
% \text{low}, & s_t(x) < 4,\\
% \text{unclear}, & s_t(x) = 4,\\
% \text{high}, & s_t(x) > 4,
% \end{cases}
% \quad t \in \{O,C,E,A,N\}.
% \]
% We map each text $x$ to a single trait label $t(x) \in \mathbf{}\{\mathcal{D,}\text{unclear}
% \}$, where $\mathcal{D}=\{\text{O$^{H}$},  \text{C$^{H}$} , \text{E$^{H}$} , \text{A$^{H}$} , \text{N$^{H}$} , 
% \text{O$^{L}$} , \text{C$^{L}$} ,\text{E$^{L}$} ,\text{A$^{L}$} ,\text{N$^{L}$}\}$.

% Specifically, we assign $t(x)=T^H$ (or $T^L$) when the annotated score $d_T(x)$ indicates a distinctly high (or low) trait intensity. If the text scores a neutral 4 across the dimensions, or if no single trait intensity is dominant, we set $t(x)=\text{unclear}$.
\vspace{-2mm}
\subsection{Cross-Domain Preference Alignment}
\vspace{-2mm}

% \begin{table}[t]
%     \caption{
%     Overall accuracy (\%) on 200 sampled \textsc{PrefEval} questions under different preference-context settings.
%     For the Mixed and Aligned settings, five preference statements are provided per question
%     (Appendix~\ref{sec:Few-shot}).
%     For Contaminated Aligned, we add two preferences from mismatched traits to the five trait-aligned preferences as noise
%     (Appendix~\ref{sec:Few-shot-noise}).
%     }
%     \label{table:experiment_prefeval}
%     \centering
%     \begin{small}
%     \setlength{\tabcolsep}{6pt}
%     \renewcommand{\arraystretch}{1.05}
%     \begin{tabular}{l|c}
%         \toprule
%         \textbf{Preference Context} &
%         \makecell{\textbf{Overall}\\\textbf{Acc. (\%)}} \\
%         \midrule
%         Zero-shot            & 25.75 \\
%         Mixed Traits         & 29.25 \\
%         Aligned Traits       & \textbf{63.00} \\
%         Contaminated Aligned & 61.75 \\
%         \bottomrule
%     \end{tabular}
%     \end{small}
% \end{table}

We then use these annotations to test whether personality can support cross-domain preference alignment. Specifically, we consider a setting in which the LLM is not given a preference that directly answers the target question. Instead, it must rely on preferences expressed in other contexts and infer answers from them. For each target question, we construct four preference-context settings (illustrated in Fig.~\ref{fig:method-overview}): \textit{(i) Zero-shot (no preference):} the LLM answers the question without any preference context; \textit{(ii) Mixed Traits:} the LLM receives question-irrelevant preferences drawn from different personality traits; \textit{(iii) Aligned Traits:} the LLM receives question-irrelevant preferences from other domains that share the same underlying personality trait as the target question; and \textit{(iv) Contaminated Aligned:} the LLM receives question-irrelevant but trait-aligned preferences together with additional unrelated or mismatched preferences.

In each setting, we sample 200 multiple-choice questions (20 for each of 10 high/low Big-Five directions). %- (more details in Sec.~\ref{subsec:setup}):
% For setups (ii) and (iii), we include five preference statements per question (see Appendix~\ref{sec:Few-shot}). For setup (iv), we include five trait-aligned preferences and two misaligned ones from other traits as noise (see Appendix~\ref{sec:Few-shot-noise}). %\salma{An example of these setups are shown in \autoref{xxxxx}.}
We evaluate preference following using accuracy. For each question $q_i$, the model selects one of four choices, producing $\hat{a}_i \in \{1,2,3,4\}$, and we compute
$
\mathrm{Acc}(\%) = \frac{1}{200}\sum_{i=1}^{200}\mathbf{1}\{\hat{a}_i = a_i\}\cdot 100\%,
$
where $\mathbf{1}\{\cdot\}$ denotes the indicator function.

%Table~\ref{table:experiment_prefeval} shows that providing trait-aligned preferences substantially improves performance, increasing accuracy to $82.50\%$, compared with $67.50\%$ for mixed preferences and $46.25\%$ in the zero-shot setting. 
We observe two main findings from the results in Tab.~\ref{table:experiment_prefeval}:
\begin{enumerate}[leftmargin=1em,topsep=0pt, partopsep=0pt, itemsep=0pt, parsep=0pt]
%     \item\textbf{Preference context improves personalization.}
% In row (i) of ~\autoref{table:experiment1}, the model receives no preference context and achieves very low accuracy. This highlights that \textsc{PACIFIC} is intentionally challenging: the correct answer cannot be reliably inferred from common sense alone and instead requires leveraging user-specific preference.

\item \textbf{Preference context improves personalization. Aligned trait preferences outperform unrelated preferences.}
% Comparing (i) and (ii), mixed-trait preference history yields some improvement over the zero-shot ($67.50\%$ vs.\ $46.25\%$), suggesting that simply providing preference history offers limited benefit when the included preferences are not aligned with the user's underlying traits. In contrast, comparing (i) and (iii), trait-aligned preferences substantially improve accuracy from $46.25\%$ to $82.5\%$. This suggests that organizing preference history by shared underlying personality traits provides substantially more informative signals for predicting the user's unseen preferences.
% Preference context helps only when trait-aligned. Mixed-trait history barely beats zero-shot ($67.50\%$ vs.\ $46.25\%$), while trait-aligned preferences lift accuracy to $82.5\%$ (Tab. \ref{table:experiment_prefeval}), showing that organizing history by shared personality traits, not mere presence of history, drives the gain of predicting the user's unseen preferences.
Preference context improves personalization when trait-aligned. In Tab. \ref{table:experiment_prefeval}, mixed-trait preference history yields only slight improvement over zero-shot ($67.50\%$ vs. $46.25\%$), showing limited benefit when preferences are not aligned with the user's underlying traits. In contrast, trait-aligned preferences substantially improve accuracy from $46.25\%$ to $82.5\%$, showing that organizing preference history by shared personality traits provides more informative signals for predicting the user's unseen preferences.

\item \textbf{Unrelated preferences introduce noise and reduce accuracy.}
Comparing (iii) and (iv), adding preferences from other traits lowers accuracy relative to using only trait-aligned preferences ($82.5\%$ vs. $72.92\%$). This indicates that irrelevant preference statements can distract the model and weaken preference-following performance.
\end{enumerate}

However, \textsc{PrefEval} is not evenly distributed across personality traits. As shown at the bottom of Fig.~\ref{fig:pacific-dataset-overview}, its examples provide highly uneven coverage across the ten high/low Big-Five directions, with several traits substantially underrepresented. This makes it difficult to determine whether observed differences across traits reflect genuine personality effects or simply differences in data coverage, preventing reliable trait-level comparisons.

\begin{table}[ht]
    \centering
    \small
    \setlength{\tabcolsep}{0.5pt}
    \vspace{-5mm}
    \caption{
Overall accuracy (\%) on \textsc{PrefEval} across preference-context setups.
Mixed and Aligned use five preferences per question (Suppl.~\ref{sec:Few-shot}); Contaminated Aligned adds two mismatched preferences as noise (Suppl.~\ref{sec:Few-shot-noise}).
}
     \label{table:experiment_prefeval}
    \begin{tabular}{c||c|c|c|c}
        \toprule
        &
        (i) Zero-shot &
        (ii) Mixed &
        (iii) Aligned &
        \makecell{(iv) Contam.Aligned} \\
        \midrule

        \textbf{Acc. (\%)} &
        46.25 &
        67.50 &
        \textbf{82.50} &
        72.92 \\

        \bottomrule
    \end{tabular}

\end{table}
\vspace{-2mm}
\section{Psychometrics-based Dataset}\label{sec:pacificdataset}
\vspace{-2mm}
To enable systematic evaluation across the full personality spectrum, we construct a psychometrics-based preference dataset covering both high and low expressions of all five Big-Five traits. 
% Rather than relying on the naturally sparse personality coverage of existing preference data, our psychometry-based dataset is explicitly designed to provide sufficient examples for studying how personality relates to preferences and whether this relationship generalizes to previously unseen decisions.
Unlike the sparse personality coverage of existing \textsc{PrefEval} preference data, our dataset is designed to study how personality relates to preferences and whether the relationship generalizes to unseen decisions.

Grounded in prior research connecting personality with decision-making and preferences~\citep{czerniawska2021values}, our pipeline generates 1,200 curated synthetic preference--query pairs spanning 20 diverse domains, including finance, entertainment, travel, and consumer electronics. As illustrated at the bottom of Fig.~\ref{fig:pacific-dataset-overview}, the dataset provides broad coverage across all ten high/low Big-Five directions.

\subsection{Dataset Construction}

\subsubsection{Generation Pipeline}

%To establish a robust benchmark with high discriminative power, we developed a synthetic generation pipeline utilizing Gemini Pro 2.5 (with details provided in \autoref{app:data cons}) to construct the dataset $\mathcal{D}$. Each instance $i \in \mathcal{D}$ is formalized as a tuple $(p_i, q_i, a_i)$, where $p_i$ is a personality-driven preference statement, $q_i$is a multiple-choice question containing a user query $u_i$ and four candidate choices $\{c_{i,k}\}_{k=1}^{4}$ and $a_i$ is the adhering ground-truth answer. Crucially, we enforce a High Violation Probability constraint during generation: $P(c_{\text{distractor}} \mid u_i) \gg P(c_{\text{distractor}} \mid p_i, u_i)$. This mathematically ensures that a generic, non-personalized LLM response will naturally contradict the user's preference, rigorously testing a model's ability to proactively apply latent constraints rather than defaulting to baseline helpfulness. 
%Each preference $p_i$ and choice $c_{i,k}$ is annotated with a 7-point Likert scale across the five personality dimensions, where 1 represents a very low and 7 represents a very high expression of the trait’s target characteristics. The preference can be explicitly or implicitly expressed through users' interactions with LLMs. An example is shown in Appendix~\ref{sec:dataset example}. Finally, to support end-to-end and implicit evaluation, each validated pair (see \autoref{sec:strategy} ) is augmented with a brief event scenario context and a 2-to-4 turns dialogue that implicitly conveys the target preference.

The construction process of our dataset is illustrated by the green boxes in Fig.~\ref{fig:pacific-dataset-overview}. We begin with the same 20 topics used in \textsc{PrefEval} and follow a similar procedure to generate preference--question--answer triples $(p_i, q_i, a_i)$. The key difference is that our generation process is explicitly anchored to psychometric personality traits. Specifically, we generate examples around each of the ten high/low Big-Five directions, ensuring that the resulting preferences cover 10 personality traits evenly.

% To make the question--answers genuinely require personalization, we further impose a high violation-probability constraint during generation:$P(c_{\text{distractor}} \mid u_i) \gg P(c_{\text{distractor}} \mid p_i, u_i).$ This constraint ensures that a generic, non-personalized LLM response will naturally contradict the user's preference, rigorously testing a model's ability to proactively apply latent constraints rather than defaulting to baseline helpfulness.

To ensure each question genuinely requires personalization, we require at least one incorrect choice $c_{\text{distractor}} \in \{c_{i,k}|k \neq a_i\}$, to satisfy $P(c_{\text{distractor}} \mid q_i) \gg P(c_{\text{distractor}} \mid q_i, p_i)$, where $P(c \mid \cdot)$ denotes how likely $c$ is to be an appropriate answer given the available information. Intuitively, $c_{\text{distractor}}$ appears plausible from the query alone, but becomes clearly inappropriate once the user's preference $p_i$ is known. This ensures that a generic, non-personalized response, conditioned on the query alone, will tend to contradict the user's preference, testing whether a model can proactively apply the latent constraint $p_i$ rather than defaulting to baseline helpfulness.

%To make the question--answers genuinely require personalization, we impose a high violation-probability constraint during generation: $P(c_{\text{distractor}} \mid q_i) \gg P(c_{\text{distractor}} \mid q_i, p_i)$, where $P(c \mid \cdot)$ denotes the probability that choice $c$ is considered an appropriate answer given the available information and $c_{\text{distractor}}$ denotes an incorrect, preference-inconsistent answer choice for question $q_i$. Intuitively, the choice $c_{\text{distractor}}$ is a plausible answer to the query $q_i$ on its own, but becomes unlikely once the user's preference $p_i$ is taken into account. This ensures that a generic, non-personalized response, conditioned on the query alone, will tend to contradict the user's preference, testing whether a model can proactively apply the latent constraint $p_i$ rather than defaulting to baseline helpfulness.
% \salma{What is u? what do you mean by $c_{distractor}$? Is this the noisy choice based on mixed trait? or are those hard negatives? FIXED}

As in \textsc{PrefEval}, we use an LLM-based judge to filter out invalid or insufficiently challenging samples. The validated preferences may be expressed either explicitly or implicitly through user--LLM interactions. To support end-to-end evaluation in more natural conversational settings, each validated example is further augmented with a brief scenario and a 2--4-turn dialogue that implicitly conveys the target preference. An example is provided in Suppl. ~\ref{sec:dataset example}. 
% \salma{Use the right reference in the suppl material not appendix FIXED}

Finally, we annotate each preference $p_i$ and each answer choice $c_{i,k}$ with one of the ten personality directions in
$\mathcal{D},$
using the same annotation procedure applied to \textsc{PrefEval}, as described in Sec.~\ref{subsec:Personality Trait Annotation}.

\subsection{Dataset Quality Control and Validation}
\label{sec:strategy}

To ensure the high fidelity and difficulty of our psychometrics-based dataset, we implemented a multi-stage validation protocol combining automated filtering, trait-matched human evaluation, and plausibility assessment as shown in Fig.~\ref{fig:pacific-dataset-overview}.

\paragraph{Automated Quality Control} To avoid generator-evaluator bias, an independent model (GPT-4o-mini) assessed all instances scalably. Only examples passing two strict rubrics were retained: (1) the ground-truth optimally aligns with the preference while the three distractors violate or ignore it, and (2) the preference logically reflects its assigned trait.

\paragraph{Human Psychometric Grounding}
To ensure ecological validity, we recruited 15 annotators from diverse professional backgrounds (e.g., PhD students, software and hardware engineers, data analysts) and ran a three-stage study:

\begin{enumerate}[leftmargin=12pt,itemsep=0.2pt]
    \item \textbf{Trait profiling.}\, Annotators completed the \cite{openpsychometrics2019ipipbffm}, based on Goldberg’s Big-Five factor markers\,\citep{goldberg1992bigfive}, to record their OCEAN profiles.
    \item \textbf{Choice validation.}\, Annotators were assigned 25 randomly sampled preference–question pairs corresponding to their own dominant trait, spanning the ten trait labels in $\mathcal{D}$. Each variant had at least five annotators. When asked to select the generated option best matching the stated preference, %because annotators were evaluating traits they personally possess,
    they demonstrated strong, intuitive agreement on trait manifestation, achieving an average accuracy of $78.22\%$ and Fleiss' $\kappa$ of $0.8599$ among users with the same assigned questions. Intuitively, this is because annotators were evaluating traits they personally possess. The GPT-4o-mini evaluator reached a Cohen's $\kappa$ of $0.9170$ against this human consensus, validating our automated pipeline.
    
    This performance of $78.22\%$ is appropriate given that each question includes carefully crafted \emph{hard-negative} distractors to prevent LLMs from exploiting surface-level keywords; the high \(\kappa\) confirms human annotators were highly consistent rather than guessing. 
    \item \textbf{Resonance check.}\, We assigned annotators preference-question pairs matched to their dominant traits. They then indicated whether they preferred the ground-truth option over the other three options. Choosing the ground truth indicated that this option offered the best alignment for the question. Ultimately, they reported a $91\%$ approval rate.
    % Annotators then indicated whether they personally identified with each synthesized preference. Despite preferences being context-dependent, they reported a $67.11\%$ approval rate. 
    % Because our dataset binarises continuous traits into high/low labels (\autoref{subsec:Dual-Strategy Annotation}), a participant may score "High" in Extraversion but lack the specific intensity required to personally resonate with an extreme High-Extraversion scenario, indicating roughly 67\% real-world resonance and confirming a strong correlation between generated preferences and real-world psychometric profiles.  
\end{enumerate}

\paragraph{Distractor Plausibility Assessment} To empirically validate our High-Violation constraint, we evaluated an LLM on the dataset queries with the user preference completely withheld. With preference withheld, an LLM scores at random guess chance ($25\%$), confirming all four options are equally attractive absent the personality constraint (Tab. \ref{table:experiment1}, zero-shot).

% The model achieved ~25\% accuracy (random chance). This confirms that without the latent personality constraint, all four options are equally attractive, ensuring models cannot succeed by relying on generic helpfulness priors.

\paragraph{Dataset Statistics} For each of the ten high/low Big-Five trait labels, we curate 120 preference--query pairs, in total 1,200 samples. As shown at the bottom of Fig.~\ref{fig:pacific-dataset-overview}, our psychometrics-based dataset provides broader and more balanced coverage across all trait directions than \textsc{PrefEval}, enabling more systematic evaluation across the full personality spectrum.

\vspace{-2mm}
\section{PACIFIC Framework}
\label{sec:framework}
\vspace{-2mm}

We revisit RQ1 (Fig.~\ref{fig:method-overview}) on our psychometrics-based dataset using the same four preference-context settings as before (Zero-shot, Mixed, Aligned, Contaminated Aligned), again \textbf{never} providing a preference that directly reveals the answer.

\subsection{Generalizability Across Psychometric Profiles}
\vspace{-2mm}
We evaluate four models: Gemma-3-4B-IT, Llama-3-8B-Instruct, Gemini-2.5-Pro, and GPT-4o-mini. Results for Gemma-3-4B-IT are reported in Tab.~\ref{table:experiment1}, while results for the remaining models are provided in Suppl.~\ref{app:exp_other_models}. We observe the same overall pattern as on \textsc{PrefEval} (Tab.~\ref{table:experiment_prefeval}): trait-aligned preference history substantially outperforms mixed-trait history, while introducing a moderate amount of mismatched preference information causes only a small degradation in performance. This replication on our psychometrics-based dataset provides further evidence that the benefit of personality-aligned preference history is not specific to the particular distribution of \textsc{PrefEval}. Notably, given clean trait-aligned context, strong models approach the ceiling (Gemini-2.5-Pro: $99.25\%$, Suppl. Tab. \ref{tab:pacific_results_gemini}), confirming the bottleneck is context construction.

More importantly, our psychometrics-based dataset's balanced coverage across personality traits allows us to examine how this effect varies across personality profiles. We evaluate all 32 personality profiles (each trait can take one of two levels (\textit{high} or \textit{low}), for a total of $2^5=32$ distinct combination profiles) and measure how much trait-aligned preferences improve performance across personas. We find that personas $(O^L, C^L, E^L, A^L, N^H)$ and $(O^L, C^L, E^L, A^L, N^L)$ achieve the highest accuracy when prompted with trait-aligned preferences (Suppl.~\ref{sec:persona analysis}). Finally, we perform 3-fold cross-validation on our psychometrics-based dataset; the results, reported in Suppl.~\ref{app:cross_validation}, show the same overall trends as Tab.~\ref{table:experiment1}.

\begin{table}[t]
\centering
\small
\setlength{\tabcolsep}{2.0pt}
\vspace{-6mm}
\caption{Overall and trait-wise accuracy (\%) across preference-context setups on our psychometrics-based dataset.}

\label{table:experiment1}
\begin{tabular}{c||c|ccccc|ccccc}
\toprule
& \textbf{Acc (\%)}
& $\mathbf{O^H}$ 
& $\mathbf{C^H}$ 
& $\mathbf{E^H}$ 
& $\mathbf{A^H}$ 
& $\mathbf{N^H}$ 
& $\mathbf{O^L}$ 
& $\mathbf{C^L}$ 
& $\mathbf{E^L}$ 
& $\mathbf{A^L}$ 
& $\mathbf{N^L}$ \\
\midrule

(i) Zero-shot
& 25.75
& 17.50 & 25.00 & 27.50 & 30.00 & 37.50
& 22.50 & 17.50 & 35.00 & 25.00 & 20.00 \\

(ii) Mixed
& 29.25
& 25.00 & 12.50 & 2.50 & 35.00 & 42.50
& 40.00 & 32.50 & 50.00 & 37.50 & 15.00 \\

(iii) Aligned
& \textbf{63.00}
& 47.50 & 42.50 & 50.00 & 67.50 & 50.00
& 80.00 & 95.00 & 85.00 & 67.50 & 45.00 \\

(iv) Contam. Aligned
& 61.75
& 52.50 & 35.00 & 35.00 & 50.00 & 60.00
& 77.50 & 87.50 & 87.50 & 85.00 & 47.50 \\

\bottomrule
\end{tabular}
\vspace{-1mm}

\end{table}

\subsection{Label-Aware Prompting Strategies}
\vspace{-2mm}
Having established in RQ1 that trait-aligned preference history can substantially improve preference following, we next ask a more practical question: \textit{how should personality information be incorporated into the LLM to best support personalization?} 

% We evaluate prompting strategies for incorporating persona information into preference-based multiple-choice QA. Given a question $q_i$ (four choices) and additional context, the model predicts $\hat{a}_i \in \{1,2,3,4\}$. Figure~\ref{fig:method-overview} summarizes the methods.

To isolate the model's ability to reason over persona traits from the complexities of information retrieval, we first assume access to the ground-truth trait labels of both the user’s question and their preference statements. %Given a question $q_i$ and candidate choices, the model predicts an answer $\hat{a}_i \in \{1,2,3,4\}$(see Figure~\ref{fig:method-overview} for an overview). 
As illustrated in RQ2 of Fig.~\ref{fig:method-overview}, we investigate three prompting strategies that expose persona traits at different levels of explicitness, ranging from direct trait labels to implicit reminders:

\textbf{A. Few-shot (Preference-only)}
% As shown in the top-middle panel of ~\autoref{fig:method-overview}, this setup corresponds to Experiment~(ii) (trait-aligned preferences) in ~\autoref{motivation}. We assume the query trait of $q_i$ is known (from the \textsc{PACIFIC} annotations) and select trait-aligned five preference statements $\{p_k\}_{k=1}^{5}$ as in-context examples (prompt template in Appendix~\ref{sec:Few-shot}). In practice, trait must be inferred from a user’s conversation history and may be noisy; mismatches can retrieve irrelevant preferences and reduce accuracy, consistent with our preliminary findings in ~\autoref{motivation}.
% \salma{An example of this method is shown in \autoref{Appendix xxxx}.}
% We formulate the prompt using five trait-aligned preference statements as in-context examples, corresponding to setup (ii) in ~\autoref{motivation}. This assumes the target query trait is perfectly matched, establishing our baseline for label-aware generation.
% We provide five trait-aligned preference statements as in-context examples, without revealing personality labels. This corresponds to the aligned-preference setting from RQ1 and serves as our baseline.
Five trait-aligned preferences are provided as in-context examples -- the aligned setting from RQ1, serving as the baseline (prompt in Suppl.~\ref{sec:Few-shot}).

%Our baseline uses five trait-aligned preference statements as in-context examples, assuming a perfectly matched target query trait. 
\textbf{B. Explicit Persona Trait Hints}
% We treat the query trait for $q_i$ as known (from the \textsc{PACIFIC} annotations) and making user's persona signal explicit in three variants. 
% We explicitly inject ground-truth trait annotations into the prompt to evaluate the model's reliance on overt psychometric signals. We test three ablations: \textit{(1) Labeled Prefs:} Appending trait labels to the preference statements. \textit{(2) Labeled Prefs + Choices:} Attaching labels to both the preferences and the candidate choices. \textit{(3) Traits Only:} Supplying only the trait labels while completely withholding the raw preference and choices text. Prompt templates can be found in Appendix~\ref{sec:Few-shot} and Appendix~\ref{sec:few-shot+persona}.
% We explicitly provide ground-truth personality annotations to test whether directly exposing the underlying trait structure further improves personalization. We consider three variants: (1) labeling the preference statements, (2) labeling both the preferences and answer choices, and (3) providing only the trait labels while withholding the preference text.
Ground-truth trait labels are exposed in three variants: labeling (1) preferences, (2) preferences and answer choices, or (3) trait labels only with preference text withheld (templates in Suppl.~\ref{sec:few-shot+persona}).

\textbf{C. Implicit Persona Reminder}
Strategy A is augmented with a lightweight instruction to consider the user's personality, testing whether encouraging persona-aware reasoning helps without explicit labels (template in Suppl.~\ref{sec:reminder}).

Together, these strategies compare how effectively trait-label-aware prompting improves LLMs' preference following.

\begin{table*}[ht]
  \centering
  \begin{small}
  \setlength{\tabcolsep}{1pt}
  \vspace{-5mm}
  \caption{Accuracy (\%) of different ways of adding persona information to preference prompting.}
  \label{tab:persona_prompting_results}
  \begin{tabular}{
    >{\raggedright\arraybackslash}p{1.6cm}
    >{\raggedright\arraybackslash}p{2.7cm}
    |c|ccccc|ccccc}
    \toprule

    \multirow{2}{*}{\textbf{Method}} &
    \multirow{2}{*}{\textbf{Variant}} &
    \multirow{2}{*}{\makecell{\textbf{Acc.}\\\textbf{(\%)}}} &
    \multicolumn{10}{c}{\textbf{Trait-wise Acc. (\%)}} \\

    \cmidrule(lr){4-13}
    & & &
    \textbf{O$^{H}$} & \textbf{C$^{H}$} & \textbf{E$^{H}$} &
    \textbf{A$^{H}$} & \textbf{N$^{H}$} &
    \textbf{O$^{L}$} & \textbf{C$^{L}$} & \textbf{E$^{L}$} &
    \textbf{A$^{L}$} & \textbf{N$^{L}$} \\

    \midrule

    \makecell[l]{A.\\Few-shot}
      & \makecell[l]{Trait-aligned\\preferences}
      & 63.00
      & 47.50 & 42.50 & 50.00 & 67.50 & 50.00
      & 80.00 & 95.00 & 85.00 & 67.50 & 45.00 \\

    \midrule

    \multirow{3}{1.6cm}{\raggedright
      B. + persona\\hints}
      & Labeled Prefs
      & \textbf{76.00}
      & 55.00 & 70.00 & 72.50 & 82.50 & 67.50
      & 87.50 & 87.50 & 90.00 & 90.00 & 57.50 \\

      & \makecell[l]{Labeled Prefs\\+ choice}
      & 57.25
      & 100.00 & 95.00 & 100.00 & 82.50 & 15.00
      & 40.00 & 55.00 & 47.50 & 17.50 & 20.00 \\

      & Traits only
      & 37.75
      & 100.00 & 75.00 & 97.50 & 67.50 & 7.50
      & 7.50 & 2.50 & 0.00 & 0.00 & 20.00 \\

    \midrule

    \makecell[l]{C. +\\Reminder}
      & \makecell[l]{Instruction-only\\(no labels)}
      & 67.00
      & 47.50 & 57.50 & 50.00 & 65.00 & 65.00
      & 82.50 & 82.50 & 90.00 & 75.00 & 55.00 \\

    \bottomrule
  \end{tabular}

  \end{small}
  \vspace{-1mm}

\end{table*}

\vspace{-2mm}
\subsection {Personality-Informed RAG (PiRAG)}
\vspace{-2mm}
In real-world conversations, personality labels are not explicitly attached to user query statements. A practical personalization system must therefore identify personality-relevant evidence directly from raw interaction history.
\textit{Without explicit personality labels, can a retriever identify personality-aligned memories that support the prediction of unseen preferences?}

Label-aware strategies establish that personality alignment helps, but ground-truth trait annotations are impractical in deployment, where labels are latent. To select trait-consistent context automatically, we propose a retrieval-augmented generation (RAG) framework.
We deploy a dual-encoder Dense Passage Retrieval (DPR) architecture~\citep{karpukhin-etal-2020-dense} to map both the user query and the candidate preference statements into a shared dense vector space. Given a query $q_i$, the system retrieves the top-k preferences based on embedding similarity to dynamically construct the LLM prompt.

%Crucially, standard semantic retrieval often fails to capture latent psychometric alignment, favoring surface-level relevance over personality consistency.
Crucially, standard semantic retrieval prioritizes topical relevance over psychometric alignment, without explicitly favoring trait-consistent evidence~\citep{salemi2024lamp,wegmann2022same}. To address this, we introduce PiRAG, a persona-aware fine-tuning step for the retriever(shown in RQ3 in Fig.~\ref{fig:method-overview}). We construct contrastive training pairs in which queries and preferences that share the same underlying OCEAN trait form positive pairs, while mismatched traits serve as negatives. This targeted fine-tuning objective pulls the representation space of same-trait representations closer together, encouraging retrieval of personality-consistent user context.
% \salma{This is a big claim. You need a citation or a realistic example. FIXED}

\vspace{-2mm}
\section{Experiments}\label{sec:evaluation}
\vspace{-2mm}
\subsection{Setup}
\label{subsec:setup}
\vspace{-2mm}
\textbf{Dataset.}
To evaluate performance across the full personality spectrum, we use our psychometrics-based dataset. We construct disjoint context and test sets as follows:

\noindent\hangindent=10pt\hangafter=1
\emph{1. Context Train Set.} Per trait, randomly draw five preference statements to form the user profile shown in the prompt.\par
\noindent\hangindent=10pt\hangafter=1
 \emph{2. Test Set.} Simultaneously, sample two \emph{non-overlapping} statements to evaluate the model.\par
\noindent\hangindent=10pt\hangafter=1
 3. Repeat steps 1–2 ten times \emph{without replacement}, producing 50 contexts, 20 tests per trait.\par
\noindent\hangindent=10pt\hangafter=1
 4. Run the entire procedure with two random seeds (200 tests/seed) and report the mean.

\textbf{Prompt format.}
We use a consistent prompt structure across methods: user preference context followed by the target query.  Unless otherwise specified, we include five preference statements in the prompt for each question. 

\textbf{LLM.}
We evaluate 4 models: two open-weight models, Gemma-3-4B-IT, Llama-3-8B-Instruct, run on the same GPU resources detailed in Suppl.~\ref{sec:hardware configuration}; and Gemini-2.5-pro and gpt-4o-mini, run via API. Gemma results are reported in Tab.~\ref{tab:persona_prompting_results}, and other models' results can be found in Suppl.~\ref{app:exp_other_models}. 

\textbf{Retrieval.}
%For retrieval-augmented prompting, 
We adopt the Dense Passage Retrieval (DPR) architecture~\citep{karpukhin-etal-2020-dense}. We use the standard DPR setup with BERT-based bi-encoders~\citep{devlin-etal-2019-bert}. Fine-tuning details are in Suppl.~\ref{app:rag}.

\subsection{Results}
\vspace{-2mm}
% \autoref{tab:persona_prompting_results} reports the accuracy across prompting variants. Overall, performance improves when the prompt provides either (i) preferences that are aligned with the target persona or (ii) simple, well-placed guidance that encourages the model to reason about persona implicitly. In contrast, explicitly adding trait labels is only helpful when those labels are reliable; noisy or weakly grounded labels can hurt.
 Overall, accuracy improves significantly when prompts include trait-aligned preferences or implicit persona guidance as shown in Tab.~\ref{tab:persona_prompting_results}, demonstrating that organizing preferences via latent traits streamlines personalization. The results of personality alignment via our \textsc{PACIFIC} framework remain consistent
across fine-tuning of both open-source small-scale and high-performance, large-scale models (Suppl.~\ref{app:exp_other_models}).

\textbf{Explicit trait labels enhance preference grounding.}
% In Table~\ref{table:experiment_prediction_accuracy} row (B), augmenting each preference with its ground-truth trait label(annotation in \textsc{PACIFIC} dataset) improves accuracy over preference-only few-shot prompting (76\% vs.\ 63\%). This suggests that explicit trait cues help the model link stated preferences to the most appropriate answer. The improvement is consistent across all ten trait settings, with trait-labeled preferences outperforming trait-aligned preferences alone in every dimension.
% Augmenting preferences with ground-truth trait labels yields the highest accuracy ($76\%$, Tab.~\ref{tab:persona_prompting_results} B), outperforming preference-only few-shot prompting ($63\%$, Tab.~\ref{tab:persona_prompting_results} A). This consistent improvement across all ten trait dimensions suggests that explicit psychometric cues help models effectively bridge stated preferences with downstream choices.
Adding ground-truth trait labels to preferences gives the highest accuracy ($76\%$; Tab.~\ref{tab:persona_prompting_results} B), improving over few-shot prompting consistently across all ten trait dimensions. This indicates that explicit psychometric cues help bridge stated preferences with downstream choices.

\textbf{Labeling answer choices triggers positivity bias.}
% When we additionally attach trait labels to the candidate answers, accuracy drops sharply (to 57.25\% in Table.~\ref{table:experiment_prediction_accuracy} row (B)).  Using only trait labels for preferences and choices performs even poorly (37.75\%). As we observe, the decline is especially pronounced for \emph{low}-trait conditions. One plausible explanation is that the model exhibits a positivity bias—preferring affirmative or “positive” choices over negative ones—which disproportionately hurts performance when the target persona signal corresponds to low-trait tendencies. This issue is explored further in ~\autoref{Negative Traits}.
% Conversely, appending trait labels to candidate answers sharply degrades performance to $57.25\%$, with a further drop to $37.75\%$ when raw text is removed entirely (Tab.~\ref{tab:persona_prompting_results} B). This decline is disproportionately driven by \emph{low}-trait conditions. We hypothesize this stems from an LLM positivity bias: models inherently favor affirmative or "positive" choices, which heavily penalizes accuracy when the target persona corresponds to low-trait tendencies. We analyze this representational collapse further in Sec.~\ref{Negative Traits}.
Conversely, labeling the candidate answers sharply degrades performance ($76\%$ -> $37.75\%$, $57.25\%$, Tab.~\ref{tab:persona_prompting_results} B), driven disproportionately by \emph{low}-trait conditions. We attribute this to an LLM positivity bias toward affirmative choices, which penalizes low-trait personas (analyzed in Sec.~\ref{Negative Traits}).

\textbf{Implicit reminders activate latent persona reasoning.}
% Replacing explicit trait labels with a brief reminder to consider the user’s persona still achieves strong overall accuracy (67\% in  ~\autoref{table:experiment_prediction_accuracy} row (C)). While this is slightly below the setting with labeled preferences, it outperforms using trait-aligned preferences alone. This suggests that a lightweight instruction can encourage persona-aware reasoning and improve preference-following, even without providing explicit trait labels.
% Replacing explicit labels with a lightweight system instruction to consider the user's persona achieves a strong $67\%$ accuracy (Tab.~\ref{tab:persona_prompting_results} C). While slightly trailing the explicit label setup, it significantly outperforms the preference-only baseline. This demonstrates that models can successfully activate latent psychometric reasoning without requiring explicit categorical annotations.
A lightweight instruction to consider the user's persona ($67\%$, Tab.~\ref{tab:persona_prompting_results} C) trails explicit labels but clearly beats the preference-only baseline, showing models can activate latent psychometric reasoning without categorical annotations.

\textbf{Contrastive fine-tuning is necessary for persona-aware retrieval.} 
% Retrieval augmentation offers a practical alternative when trait annotations for both queries and preferences are absent—a common real-world scenario. As shown in Tab.~\ref{tab:rag_results} (row 1), a pretrained retriever yields 30.25\% accuracy—below label-based setups (Tab.~\ref{tab:persona_prompting_results}) but already better than using no preferences or mixed preferences in Tab.~\ref{table:experiment1}. This gap is largely because the pretrained retriever favors semantic similarity over persona consistency. Our Fine-tuning of DPR (detailed implementations in (Suppl.~\ref{app:rag}) and trait-wise accuracy in Tab.~\ref{tab:rag}) improves performance to 43\% (Tab.~\ref{tab:rag_results} (row 2)), suggesting the retriever learns to retrieve trait-aligned preferences. Although retrieval still lags behind Few-shot and Reminder prompting, as the retriever has no direct information about the trait annotations, it still highlights the necessity of shaping dense retrieval spaces around personality congruence. Due to time constraints, we only evaluated a DPR-style retriever; we expect that more persona-aware retrieval methods could further close this gap.
When trait annotations are absent — the common real-world case — a pretrained DPR retriever yields only $30.25\%$ (Tab.~\ref{tab:rag_results}), since it favors semantic similarity over persona consistency, though this already beats the no-preference and mixed-preference settings (Tab.~\ref{table:experiment1}). Persona-aware contrastive fine-tuning raises accuracy to $43\%$ (Tab.~\ref{tab:rag_results}; implementation details in Suppl.~\ref{app:rag} and trait-wise accuracy in Tab.~\ref{tab:rag}), showing the retriever learns to surface trait-aligned preferences. Ground-truth-label prompting remains higher, but requires trait annotations that are impractical at deployment; PiRAG is the label-free counterpart. The $43\%$ is a conservative floor from a deliberately simple DPR retriever; stronger encoders should raise it.
Notably, expanding mixed-trait context from 5 to 25 preferences slightly reduces accuracy (from $61.5\%$ to $59.5\%$; Suppl.~\ref{app:exp_other_models}, Tab.~\ref{tab:pacific_results_gemini}), indicating more raw history is not better. We discuss more in Suppl.~\ref{sec: memory discussion}. We further repeat the same experiments on \textsc{PrefEval} and observe consistent trends (Suppl.~Tab.~\ref{tab:prefeval_results}), further supporting our findings.

%It still trails prompting with ground-truth labels, but confirms the value of shaping the retrieval space around personality congruence. We evaluated only a DPR-style retriever; stronger persona-aware methods could close this gap further.

% \salma{Can we write this sentence here instead of the conclusion? Yes i think its ok. FIXED}

% Our results also suggest that agentic memory should favor semantic abstraction over lossless storage: 
% adding more raw preference history hurts rather than helps (expanding mixed-trait context from 5 to 25 preferences reduces accuracy from $61.5\%$ to $59.5\%$; Suppl.~\ref{app:exp_other_models}, Tab.~\ref{tab:pacific_results_gemini}), so personality traits can thus serve as a high-signal compression layer for predicting user intent. We discuss more for agentic memory in Suppl.~\ref{sec: memory discussion}.
% Notably, expanding mixed-trait context from 5 to 25 preferences slightly reduces accuracy (from $61.5\%$ to $59.5\%$; Suppl.~\ref{app:exp_other_models}, Tab.~\ref{tab:pacific_results_gemini}), indicating more raw history is not better. We discuss more for agentic memory in Suppl.~\ref{sec: memory discussion}.

\begin{table}[h]
  \centering
  \small
  \setlength{\tabcolsep}{4pt}
  \renewcommand{\arraystretch}{1.0}
\vspace{-3mm}
  \caption{Overall accuracy (\%) using a pretrained versus persona-aware fine-tuned retriever.}
  \label{tab:rag_results}
  \begin{tabular}{c|cc}
    \toprule
    \textbf{RAG} & 
    \makecell{Pretrained Retriever} &
    \makecell{Fine-tuned Retriever(PiRAG)} \\
    \midrule

    \textbf{Acc. (\%)} & 30.25 & 43.00 \\

    \bottomrule
  \end{tabular}
  \vspace{-3mm}
\end{table}

\vspace{-2mm}
\begin{table*}[ht]
 \caption{Persona-trait prediction accuracy (\%) on 200 samples using Gemma-3-4B-IT.}
  \label{table:experiment_prediction_accuracy}
  \centering
  \begin{small}
  \setlength{\tabcolsep}{3pt}
  \begin{tabular}{lccccccccccc}
    \toprule
    \textbf{Input type} &
    \makecell{\textbf{Overall}\\\textbf{Acc. (\%)}} &
    \multicolumn{10}{c}{\textbf{Trait-wise Acc. (\%)}} \\
    \cmidrule(lr){3-12}
    & &
    \textbf{O$^{H}$} & \textbf{C$^{H}$} & \textbf{E$^{H}$} & \textbf{A$^{H}$} & \textbf{N$^{H}$} &
\textbf{O$^{L}$} & \textbf{C$^{L}$} & \textbf{E$^{L}$} & \textbf{A$^{L}$} & \textbf{N$^{L}$} \\
    \midrule
    Preferences & 54.5 &100 &100 &95 &100 &100 &5 &0 &0 &5 &40 \\
    Choices     & 85.69 &91.88 &63.75 &87.5 &91.25 &75 &87.5 &85.63 &91.25 &91.88 &91.25 \\
    \bottomrule
  \end{tabular}
  \end{small}
  \vskip -0.2in
\end{table*}

\subsection{Social Desirability Bias}
\vspace{-2mm}
\label{Negative Traits}

To succeed in realistic, end-to-end conversational settings, models must autonomously infer a user's latent personality from their interaction history. We evaluate this by prompting the LLM to predict OCEAN traits directly from either stated preferences or candidate choices. Tab.~\ref{table:experiment_prediction_accuracy} reveals a counterintuitive discrepancy: trait prediction from abstract answer choices ($85.69\%$) significantly outperforms prediction from explicit user preferences ($54.50\%$).

A trait-wise breakdown exposes the root cause: while the model perfectly identifies "High" traits from preferences, its accuracy collapses (0–40\%) on "Low" dimensions. We attribute this to Social Desirability Bias induced by Reinforcement Learning from Human Feedback (RLHF), which can conflate psychometric "low" scores (e.g., a preference for routine over openness) with normatively negative behavior or safety violations\citep{yi2025too,salecha2024large}. 
As a result, models may suppress such signals when judging users. This bias is context-dependent: accuracy recovers when evaluating abstract choices rather than users directly. Consequently, it can bottleneck personalization by favoring socially desirable over preference-aligned recommendations.

\vspace{-3mm}
\section{Related Work}\label{sec:related}
\vspace{-2mm}
\textbf{Personality and Human Preferences.} The correlation between the Big Five personality traits and human preferences is well-documented in psychological literature, particularly regarding aesthetic and entertainment preferences \citep{rentfrow2003re, rentfrow2011listening, cantador2013relating}. 
%These findings motivate our dataset assumptions: personality can act as a latent prior shaping observable choices. 
Building on these psychological findings, researchers have developed Personality-Aware Recommendation Systems \citep{hu2011enhancing, ning2019personet}. Treating personality as a latent prior for observable preferences motivates our dataset and framework.

\textbf{Personality, Preference and Large Language Models.} Recent work has increasingly focused on the intersection of personality psychology and Large Language Models (LLMs), shifting from simple instruction following to complex persona simulation. Early research focused on ``inducing'' specific personality traits into LLMs to make them more human-like \citep{jiang2023evaluating, li-etal-2025-big5}. While these works focus on the model's personality, a critical, complementary challenge is extracting the latent knowledge from raw interactions and aligning models with the user's personality. Recent frameworks like Proxona \citep{proxona} address this by distilling audience traits from raw comments into dimensions and values to cluster interactive personas. To effectively serve diverse users, models must understand how latent traits drive behavior, yet no standardized benchmark currently exists to evaluate this reasoning capability. The landscape of existing datasets is primarily dominated by tasks that test explicit adherence rather than psychometric understanding. Prior work \citep{zhao2025llms} focuses on logical constraint satisfaction, and work \citep{jiang2025know} targets dynamic memory retrieval; both lack explicit psychometric grounding. In contrast, our dataset and framework are the first to study the causal link between latent Big Five personality traits and downstream preferences, shifting the evaluation focus from simple rule adherence or fact recall to psychologically plausible, trait-driven preference prediction.

\section{Conclusion}\label{sec:discuss}

In conclusion, we introduce PACIFIC, a psychometrics-based framework demonstrating that leveraging latent personality traits enhances LLM preference alignment accuracy, improving label-free retrieval accuracy by $43\%$ relative ($30{\to}43\%$) over standard semantic retrieval, while clean trait-aligned context approaches ceiling.

\newpage

% \section*{Acknowledgements}

% This work is supported by the U.S. National Science Foundation (NSF) under grant number 2339266 and is partially supported by the UCI ProperAI Institute, an Engineering+Society Institute funded as part of a generous gift from Susan and Henry Samueli.

% Authors are \textbf{required} to include a statement of the potential broaderimpact of their work, including its ethical aspects and future societal
% consequences. This statement should be in an unnumbered section at the end of
% the paper (co-located with Acknowledgements -- the two may appear in either
% order, but both must be before References), and does not count toward the paper
% page limit. In many cases, where the ethical impacts and expected societal
% implications are those that are well established when advancing the field of
% Machine Learning, substantial discussion is not required, and a simple
% statement such as the following will suffice:

% ``This paper presents work whose goal is to advance the field of Machine
% Learning. There are many potential societal consequences of our work, none
% which we feel must be specifically highlighted here.''

% The above statement can be used verbatim in such cases, but we encourage
% authors to think about whether there is content which does warrant further
% discussion, as this statement will be apparent if the paper is later flagged
% for ethics review.

% \section*{Author Contributions}

\section*{Acknowledgments}
This work is supported by the U.S. National Science Foundation (NSF) under grant number 2339266 and is partially supported by the UCI ProperAI Institute, an Engineering+Society Institute funded as part of a generous gift from Susan and Henry Samueli.

% \section*{Ethics Statement}
% Authors can add an optional ethics statement to the paper. 
% For papers that touch on ethical issues, this section will be evaluated as part of the review process. The ethics statement should come at the end of the paper. It does not count toward the page limit, but should not be more than 1 page. 

\bibliography{colm2026_conference}
\bibliographystyle{colm2026_conference}

\appendix
%%%%%%%%%%%%%%%%%%%%%%%%%%%%%%%%%%%%%%%%%%%%%%%%%%%%%%%%%%%%%%%%%%%%%%%%%%%%%%%
%%%%%%%%%%%%%%%%%%%%%%%%%%%%%%%%%%%%%%%%%%%%%%%%%%%%%%%%%%%%%%%%%%%%%%%%%%%%%%%
% APPENDIX
%%%%%%%%%%%%%%%%%%%%%%%%%%%%%%%%%%%%%%%%%%%%%%%%%%%%%%%%%%%%%%%%%%%%%%%%%%%%%%%
%%%%%%%%%%%%%%%%%%%%%%%%%%%%%%%%%%%%%%%%%%%%%%%%%%%%%%%%%%%%%%%%%%%%%%%%%%%%%%%
\newpage
\appendix
\onecolumn

\section{Dataset}
\subsection{Dataset Example}
\label{sec:dataset example}
We provide an example from our Psychometrics-based Dataset. Tab.~\ref{tab:pacific_pref_example} presents the annotated preference statement, and Tab.~\ref{tab:pacific_choice_example} lists the four corresponding answer choices with their labels and OCEAN annotations. 

\subsection{Annotation Criteria}
\label{app:criteria}

% \vspace{-2mm}
\subparagraph{Scoring Criteria for O, C, E, A:} To operationalize this, we define the annotation scale based on the \textit{spectrum} of the trait characteristics present in the preference statement:
\begin{itemize}[nosep, leftmargin=*]
    \item \textit{Score 1-3 (Low Trait Expression)}: The preference reflects the characteristics associated with the low end of the trait spectrum (e.g., Routine/Tradition for Openness). 
    \item \textit{Score 4 (Neutral)}: The preference shows no strong directional alignment with the trait.
    \item \textit{Score 5-7 (High Trait Expression): } The preference reflects the characteristics associated with the high end of the trait spectrum (e.g., Novelty/Complexity for Openness).
\end{itemize}

% \vspace{-2mm}
\subparagraph{Scoring Criteria for N:} To operationalize this, we defined the annotation scale for Neuroticism based on the provision of \textit{Safety vs. Efficiency}:
\begin{itemize}[nosep, leftmargin=*]
    \item \textit{Score 1-3 (High Efficiency / High Risk):} The preference prioritizes performance, efficiency, brutal truths, and high stakes. (Target: Low-N Users). 
    \item \textit{Score 4 (Neutral)}: A balance between safety and performance.
    \item \textit{Score 5-7 (High Safety / High Comfort):} The preference prioritizes guarantees, emotional regulation, and risk avoidance. (Target: High-N Users).
\end{itemize}

\begin{table*}[p]
\centering
\small
\setlength{\tabcolsep}{3pt}
\caption{\textsc{PACIFIC} entry example High-E (Preference annotation).}
\label{tab:pacific_pref_example}

\begin{tabular}{@{}p{0.25\textwidth} p{0.06\textwidth} p{0.06\textwidth} p{0.60\textwidth}@{}}
\toprule
\textbf{Field} & \textbf{Trait} & \textbf{Value} & \textbf{Text} \\
\midrule

preference\_statement & -- & -- &
I only enjoy live music events where I can be in a general admission pit or a standing-room-only section close to the action.\\

related\_to\_personality & -- & 1 & --\\
\midrule

score & O & 6 & --\\
confidence & O & 0.6 & --\\
explanation & O & -- &
The user's desire to be 'close to the action' in a high-intensity environment like a general admission pit suggests a preference for new and stimulating experiences, which aligns with the 'Adventurousness' facet of high Openness.\\
confidence\_explanation & O & -- &
The preference for an intense, non-traditional concert experience is a good indicator of adventurousness. However, it's a specific context and may not generalize to other areas of life.\\
\addlinespace[4pt]

score & C & 4 & --\\
confidence & C & 0.1 & --\\
explanation & C & -- &
The statement provides no information about the user's level of organization, discipline, or reliability. The preference for a pit could be seen as either spontaneous (low) or requiring planning (high), making it impossible to assess.\\
confidence\_explanation & C & -- &
There is no evidence in the provided text to support a score for Conscientiousness in either direction.\\
\addlinespace[4pt]

score & E & 7 & --\\
confidence & E & 0.9 & --\\
explanation & E & -- &
The user's preference for a general admission pit, a crowded, high-energy, and socially dense environment, strongly indicates a high need for stimulation and excitement, which are core facets of Extraversion.\\
confidence\_explanation & E & -- &
The statement describes a classic preference for an environment sought by individuals high in Extraversion, particularly the 'Excitement-Seeking' and 'Gregariousness' facets. The use of 'only' strengthens this signal.\\
\addlinespace[4pt]

score & A & 4 & --\\
confidence & A & 0.1 & --\\
explanation & A & -- &
The statement describes a preferred environment but does not detail the user's interpersonal style (e.g., cooperative vs. competitive) within that environment. Therefore, agreeableness cannot be assessed.\\
confidence\_explanation & A & -- &
There is no evidence in the provided text to support a score for Agreeableness in either direction.\\
\addlinespace[4pt]

score & N & 1 & --\\
confidence & N & 0.8 & --\\
explanation & N & -- &
Choosing a general admission pit, an environment with physical discomfort and potential risks (crowds, pushing), over a safer, seated option demonstrates a very low need for comfort and safety, prioritizing the intense experience instead.\\
confidence\_explanation & N & -- &
The preference is a clear and direct trade-off, sacrificing physical safety and comfort for a more stimulating experience. This provides strong evidence for a low score on this specific scale.\\

\bottomrule
\end{tabular}
\end{table*}

\begin{table*}[p]
\centering
\small
\setlength{\tabcolsep}{0.8pt}
\caption{\textsc{PACIFIC} entry example High-E (Choice annotations).}
\label{tab:pacific_choice_example}

\begin{tabular}{@{}p{0.25\textwidth} p{0.05\textwidth} p{0.2\textwidth} p{0.5\textwidth}@{}}
\toprule
\textbf{Choice} & \textbf{Field} & \textbf{Trait Score(Confidence)} & \textbf{Text} \\
\midrule

\multicolumn{4}{@{}p{\textwidth}@{}}{\textbf{Conversation query:} What's the best way for me to get tickets for the upcoming city music festival?}\\

\multicolumn{4}{@{}l@{}}{\textbf{Choice 0}}\\
text & -- & -- &
Prioritize getting the 'Front Stage Pit' pass to be right in the middle of the crowd and energy.\\
related\_to\_personality & -- & 1 & --\\
score/conf & O & 5 (0.6) & explanation:The user seeks a high-intensity sensory experience by wanting...\\
score/conf & C & 4 (0.2) & explanation: The statement shows a clear goal ('Prioritize getting ...\\
score/conf & E & 7 (1.0) & explanation: The user's explicit desire to be 'right in...\\
score/conf & A & 4 (0.1) & explanation: The choice is focused on a personal sensory goal and does not...\\
score/conf & N & 1 (0.9) & explanation: The choice represents a clear trade-off, sacrificing ...\\

\midrule
\multicolumn{4}{@{}l@{}}{\textbf{Choice 1}}\\
text & -- & -- &Secure a reserved seat in the grandstand for a comfortable and clear, but distant, view.\\
related\_to\_personality & -- & 1 & --\\
score/conf & O & 3 (0.6) & explanation: The user chose a practical and conventional option (a reserved seat) ...\\
score/conf & C & 6 (0.8) & explanation: The action to 'secure a reserved seat' implies planning...\\
score/conf & E & 2 (0.8) & explanation: The user opted for a 'distant' view from a grandstand,...\\
score/conf & A & 4 (0.1) & explanation: The choice of seating is a personal preference for comfort and view; ...\\
score/conf & N & 7 (0.9) & explanation: Choosing a 'reserved' and 'comfortable' seat is a risk-averse ...\\

\midrule
\multicolumn{4}{@{}l@{}}{\textbf{Choice 2}}\\
text & -- & -- & Look for tickets in the upper levels, as they are often cheaper and far less crowded.\\
related\_to\_personality & -- & 1 & --\\
score/conf & O & 3 (0.6) & explanation: The user prioritizes practical concerns like cost ('cheaper') and...\\
score/conf & C & 6 (0.7) & explanation: The user demonstrates cautiousness and deliberation...\\
score/conf & E & 2 (0.9) & explanation: The user's explicit preference for a 'far less crowded'...\\
score/conf & A & 4 (0.1) & explanation: The choice to seek cheaper, less crowded ...\\
score/conf & N & 6 (0.8) & explanation: The user prioritizes comfort and the avoidance of potential...\\

\midrule
\multicolumn{4}{@{}l@{}}{\textbf{Choice 3}}\\
text & -- & -- & Opt for a VIP package that includes a private, quiet viewing lounge away from the masses.\\
related\_to\_personality & -- & 1 & --\\
score/conf & O & 4 (0.3) & explanation: The user is attending a music festival, which can be a novel ...\\
score/conf & C & 6 (0.6) & explanation: Opting for a VIP package implies planning and a desire ...\\
score/conf & E & 2 (0.9) & explanation: The user's explicit desire for a 'private, quiet' space 'away ...\\
score/conf & A & 4 (0.2) & explanation: The statement does not provide clear evidence regarding the user's...\\
score/conf & N & 7 (0.8) & explanation: The choice prioritizes a highly controlled, comfortable, ...\\

\bottomrule
\end{tabular}
\end{table*}

\subsection{Trait Distribution}
\label{app:trait dist}

\begin{figure}[!h]
    \centering
    \includegraphics[width=0.8\linewidth]{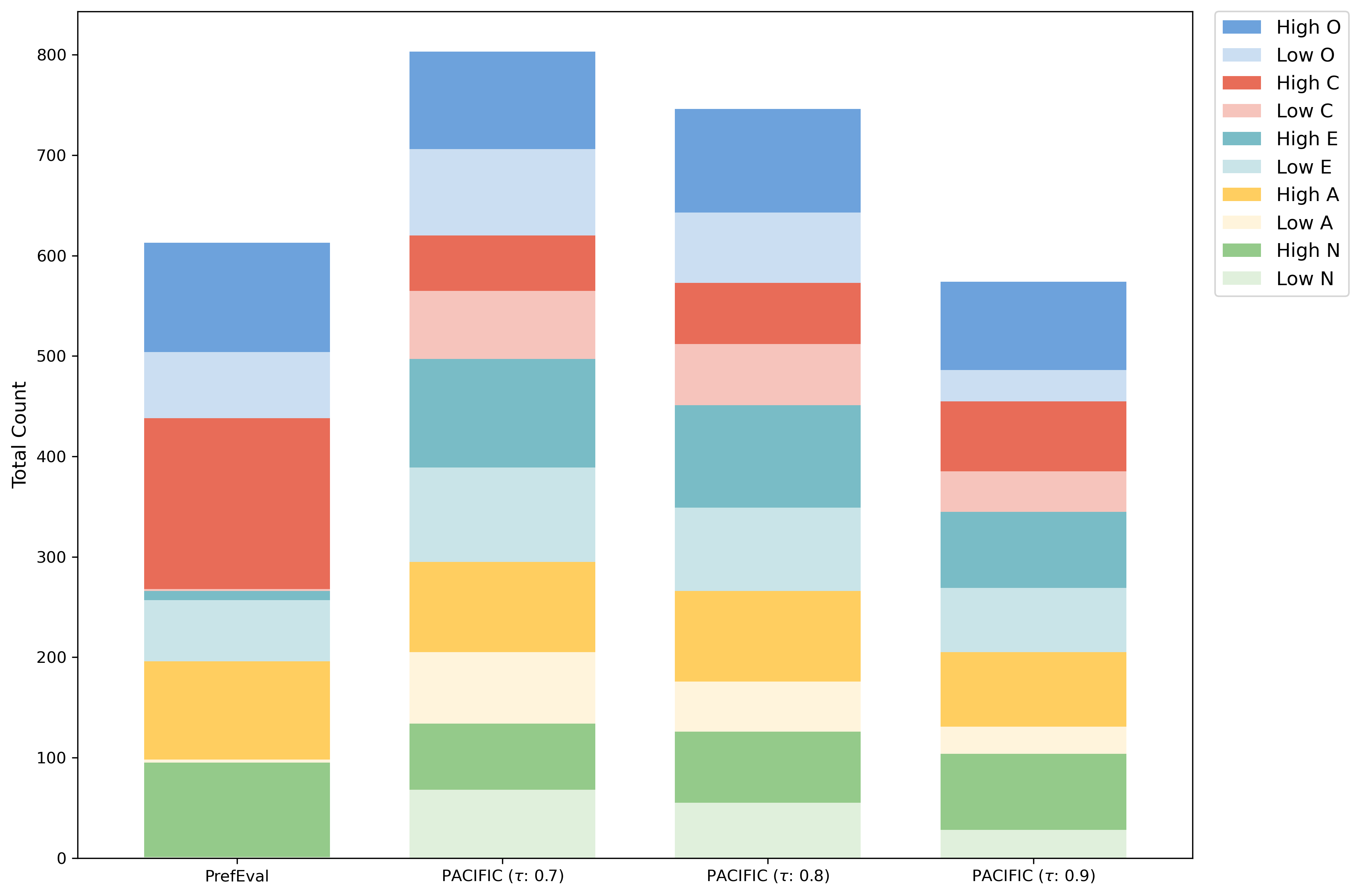}
    \caption{\textbf{Trait distribution overview:} PrefEval exhibits a skewed trait distribution, with limited examples for Low-C, High-E, Low-A, and Low-N. In total, PrefEval has 613 personality-driven preferences out of 1000 preferences with confidence of $\tau = 0.7$. \textsc{PACIFIC} yields 803, 746, 574 personality-driven preferences out of 1200 preferences with confidence of $\tau = 0.7, 0.8, 0.9$ respectively. }
    \label{fig:trait dist}
    \vspace{-5mm}
\end{figure}

 The distribution of each trait in \textsc{PrefEval} and \textsc{PACIFIC}(with $\tau = 0.7, 0.8, 0.9$, respectively) is shown in Fig. \ref{fig:trait dist}. \textsc{PACIFIC} provides complete personality-driven preferences, while PrefEval is severely skewed with few preferences in Low-C, High-E, Low-A, and Low-N. 

\vspace{-2mm}

\vspace{-2mm}
\subsection{Data construction}
\label{app:data cons}
% \vspace{+3cm}
To establish a robust benchmark for personality-aware preference alignment, we developed a two-stage pipeline utilizing the Gemini Pro 2.5 API. Our methodology prioritizes high discriminative power by focusing on scenarios where generic model behavior is insufficient.

\textbf{Personality-Driven Scenario Generation.}
The first step focused on synthesizing challenging interaction scenarios rooted in distinct personality profiles. For each instance in the dataset, we generated a structured tuple consisting of:
\begin{itemize}[noitemsep, leftmargin=*,topsep=0pt]
\item Personality-Driven Preference: A specific constraint or desire derived from a user persona.
\item Preference-Sensitive Query: A user question designed specifically to test the system's adherence to the stated preference.
\item Difficulty Rationale: A concise explanation detailing why this query presents a challenge for standard models.
\item Candidate Responses: A set of four response choices, where exactly one option aligns with the user’s preference and personality, while the remaining three serve as plausible but misaligned distractors.
\end{itemize}

A critical constraint in this process was the enforcement of a "High Violation" criterion. We prompted the model to generate queries where a generic or "safe" response would be highly likely to violate the specific user preference. This ensures that the dataset effectively filters for models capable of nuanced personalization rather than generic helpfulness. The prompt is shown in Prompt \ref{prompt: scenario} and provided in full in the supplemental material.

We also employ a Confidence Filtering Protocol to derive a high-quality evaluation subset. Let a data sample be defined as S=$\{p_i, q_i, (c_\text{correct}, c_\text{distractor})\}$, where $p$ is the preference statement, $c_\text{correct}$ is the ground-truth choice, and $c_\text{distractor}$ are the incorrect options. We adopt a confidence threshold $\tau=0.7$ and apply the following inclusion criteria:

\begin{itemize}[nosep, leftmargin=*,topsep=0pt]
\item \textbf{Preference Validity:} The preference statement $p$ must reflect the target personality trait (High or Low) with a model confidence score $\tau$. 

\item \textbf{Answer Congruence:} The correct choice $c_\text{correct}$ must not only be semantically valid but must also explicitly exhibit the same trait polarity as $p$ with confidence $\tau$. This ensures the ``correctness'' is driven by personality alignment, not just general common sense. 

\item \textbf{Distractor Separation:} To ensure discriminative power, all distractors must fail to meet the target criteria. A distractor is deemed valid only if it either reflects the opposing trait polarity or lacks sufficient confidence (conf$<\tau$) to be confused with a trait-aligned choice. 

\end{itemize}

While the total number of eligible preferences varies with different thresholds, applying a threshold of $\tau$=0.7 yields $803$ different preferences and a balanced trait distribution. We set $\tau$=0.7 to filter out uncertain pairs and avoid possible LLM hallucination. Visualizations of trait distributions under other $\tau$ values with \textsc{PrefEval} comparison are detailed in Suppl. ~\ref{app:trait dist}.

\begin{floatquote}[h]
\begin{quote}
\noindent
\scriptsize
\begin{center}
\fbox{%
  \parbox{0.95\linewidth}{%
  \vspace{0.6em}
  \texttt{You are a helpful assistant. You are helping a user create scenarios to evaluate if an AI assistant properly considers the user's stated preferences.}
    \texttt{You will generate:}\\[0.5em]  
    \texttt{Phase 1: The Scenario} \\
    \texttt{1. **Personality-driven preference**:...} \\
    \texttt{2. **Question**: ...} \\
    \texttt{3. **Explanation**: ...} \\

    \texttt{<OCEAN definitions>}\\

    \texttt{Rubric for generating the scenarios:}\\
    \texttt{Please generate such preference question pairs with high Violation probability:}
    \texttt{<High violation definition>}\\
    \texttt{<Hign violation examples>}\\
    \texttt{<Constraints>}\\

    \texttt{Phase 2: The Options} \\
    \texttt{Given the generated personality-driven preference, question and short explanation from Phase 1.} \\
    \texttt{Think of exactly 4 possible recommendation options to answer this question in the 'options' list. }\\
    \texttt{<Dual-Strategy Annotation>}\\

    \texttt{Generate exactly 4 options in the specific order below.}\\
    \texttt{* **Option 1 (First Item in list)**: The **ADHERING** response. ...}\\
    \texttt{* **Options 2, 3, 4 (Remaining Items)**: The **VIOLATING** responses. ...}

  \vspace{0.6em}
  }%
}
\end{center}
\end{quote} 
\caption{Scenario generation prompt}
\label{prompt: scenario}
\end{floatquote}

\begin{floatquote}[h]
\begin{quote}
\noindent
\scriptsize
\begin{center}
\fbox{%
  \parbox{0.95\linewidth}{%
  \vspace{0.6em}
  \texttt{You are a personality assessment expert. Analyze the following user preference statement or user behavior and predict their OCEAN (Big Five) personality trait scores.}\\
    \texttt{Input Data}\\[0.5em]  
    \texttt{<Preference or Question + 1 choice>} \\
    \texttt{Task:} \\
    \texttt{Assess the user's personality traits using the OCEAN model based on .....} \\
    \texttt{<OCEAN Personality Traits>} \\

    \texttt{<OCEAN definitions>}\\

    \texttt{Assessment Instructions}\\
    \texttt{Step 1: Relevance Check
}
    \texttt{Determine if the preference statement is related to personality traits. ...}\\
    \texttt{Step 2: Trait Scoring}\\
    \texttt{<**Trait Score** (1-7 scale integer)> }\\
    \texttt{Step 3: Assessment Guidelines}\\

    \texttt{<Guidelines>} \\

  \vspace{0.6em}
  }%
}
\end{center}
\end{quote} 
\caption{personality annotation prompt}
\label{prompt: annotation}
\end{floatquote}

\textbf{Dual-Strategy Annotation}
To guarantee the quality and interpretability of the dataset, we implemented a dual-strategy annotation scheme. This process involved assessing both the latent user preference and the manifest content of each candidate response against the Big Five (OCEAN) personality traits. Prompt can be found in Prompt \ref{prompt: annotation}. The annotation criteria details can be found in Sec.\ref{sec:strategy}.

\newpage
\subsection{Experiment result in Prefeval}\label{sec: prefeval experiment}
We report results from the same experiments in the PrefEval~\cite{zhao2025llms} dataset in Tab.~\ref{tab:prefeval_results}.

\begin{table*}[t]
  \caption{Accuracy (\%) for different ways of incorporating persona hints into preference prompting from PrefEval Dataset~\cite{zhao2025llms}.}
  \label{tab:prefeval_results}
  \centering
  \begin{small}
  \setlength{\tabcolsep}{1.5pt}
  \renewcommand{\arraystretch}{1}
  \begin{tabular}{p{2.2cm} p{3cm}|c|ccccc|ccccc}
    \toprule
    \textbf{Method} & \textbf{Variant} &
    \makecell{\textbf{Overall}\\\textbf{Acc. (\%)}} &
    \multicolumn{10}{c}{\textbf{Trait-wise Acc. (\%)}} \\
    \cmidrule(lr){4-13}
    & & &
    \textbf{O$^{H}$} & \textbf{C$^{H}$} & \textbf{E$^{H}$} & \textbf{A$^{H}$} & \textbf{N$^{H}$} &
    \textbf{O$^{L}$} & \textbf{C$^{L}$} & \textbf{E$^{L}$} & \textbf{A$^{L}$} & \textbf{N$^{L}$} \\
    \midrule
    A. Few-shot &
      \makecell[l]{Trait-aligned\\preferences}
      & 82.50 & 82.50 & 85.00 & -- & 92.50 & 75.00 & 67.50 & -- & 92.50 & -- & -- \\
    \midrule
    \multirow{3}{*}{\makecell{B. Few-shot  + \\persona hints}}
      & \makecell[l]{Pref.\ + pref.\ trait\\(GT)}
      & \textbf{82.50} & 85.00 & 80.00 & -- & 95.00 & 67.50 & 70.00 & -- & 97.50 & -- & -- \\
      & \makecell[l]{Pref.\ + pref.\ \& choice\\traits (GT)}
      & 75.83 & 70.00 & 80.00 & -- & 95.00 & 60.00 & 60.00 & -- & 90.00 & -- & -- \\
      & \makecell[l]{Traits only\\(pref.\ \& choices) (GT)}
      & 48.33 & 45.00 & 65.00 & -- & 45.00 & 40.00 & 25.00 & -- & 70.00 & -- & -- \\
    \midrule
    C. Reminder &
      \makecell[l]{Instruction-only\\(no labels)}
      & 83.75 & 85.00 & 82.50 & -- & 92.50 & 72.50 & 70.00 & -- & 97.50 & -- & -- \\
    \midrule
    D. RAG &
      \makecell[l]{Pre-trained retriever}
      & 74.59 & 65.00 & 65.00 & -- & 92.50 & 65.00 & 67.50 & -- & 92.50 & -- & -- \\
      & \makecell[l]{Fine-tuned retriever}
      & 80.84 & 85.00 & 62.50 & -- & 85.00 & 87.50 & 70.00 & -- & 95.00 & -- & -- \\
    \bottomrule
  \end{tabular}
  \end{small}
  \vskip -0.1in
\end{table*}
\subsection{Topics Covered in our psychometrics-based dataset}
\label{sec:topics}
Topics covered in \textsc{PACIFIC} dataset are listed in Table.~\ref{tab:all_topics}.
\begin{table*}[t]
\centering
\small
\setlength{\tabcolsep}{5pt}
\renewcommand{\arraystretch}{1.05}
\caption{Topic taxonomy and example subtopics.}
\label{tab:all_topics}
\begin{tabular}{p{0.22\textwidth} p{0.75\textwidth}}
\toprule
\textbf{Topic} & \textbf{Description (examples)} \\
\midrule
Home\_Cooking & Recipe modifications, kitchen equipment, meal planning, cooking techniques, food storage \\
Personal\_Finance & Bill management, credit score, daily budgeting, banking issues, savings tips \\
Home\_Maintenance & Appliance repairs, plumbing issues, cleaning methods, power problems, HVAC care \\
Family\_Care & Child development, elderly support, family activities, parenting tips, work--life balance \\
Digital\_Services & App troubleshooting, account security, subscription management, device setup, software updates \\
Weather\_Planning & Daily forecasts, event planning, storm preparation, seasonal activities, travel weather \\
Personal\_Documents & ID renewal, document filing, form completion, legal paperwork, record keeping \\
Shopping\_Assistance & Price comparison, product reviews, warranty info, return policies, discount finding \\
Time\_Management & Schedule planning, deadline tracking, calendar organization, task prioritization, routine building \\
Communication\_Help & Email writing, message drafting, call scripts, meeting planning, social media posts \\
Medical\_Care & Symptom checking, medication info, appointment booking, insurance questions, health records \\
Entertainment\_Planning & Event tickets, party planning, holiday activities, group gatherings, weekend ideas \\
Personal\_Growth & Habit formation, goal setting, self-improvement, skill development, career planning \\
Local\_Services & Business hours, service booking, location finding, price inquiries, review checking \\
Relationship\_Advice & Communication tips, conflict resolution, dating guidance, friend issues, family dynamics \\
Smart\_Home & Device control, automation setup, network issues, energy management, security settings \\
Personal\_Safety & Emergency plans, safety precautions, security tips, first aid help, risk assessment \\
Moving\_Relocation & Planning timeline, service booking, address changes, packing tips, set up utilities \\
Gift\_Giving & Gift ideas, occasion planning, budget options, wrapping tips, delivery tracking \\
Seasonal\_Tasks & Holiday planning, weather preparation, wardrobe changes, decoration ideas, activity planning \\
\bottomrule
\end{tabular}
\end{table*}

\subsection{Experimental Results on PACIFIC with Other Models}
\label{app:exp_other_models}
We evaluate different persona prompting strategies on PACIFIC across multiple models. The results for Llama-3-8B-Instruct, gemini-2.5-pro, and gpt-4o-mini are reported in Tables~\ref{tab:pacific_results_llama}, \ref{tab:pacific_results_gemini}, and \ref{tab:pacific_results_openai}, respectively.

\begin{table*}[t]
\caption{Accuracy (\%) of Llama-3-8B-Instruct under different strategies for incorporating persona hints into preference prompting on PACIFIC.}
\label{tab:pacific_results_llama}
\centering
\begin{small}
\setlength{\tabcolsep}{2pt}
\renewcommand{\arraystretch}{1.05}
\begin{tabular}{p{2cm} p{3cm}| c |ccccc|ccccc}
\toprule
\textbf{Method} & \textbf{Variant} &
\makecell{\textbf{Overall}\\\textbf{Acc. (\%)}} &
\multicolumn{10}{c}{\textbf{Trait-wise Acc. (\%)}} \\
\cmidrule(lr){4-13}
& & &
\textbf{O$^{H}$} & \textbf{C$^{H}$} & \textbf{E$^{H}$} & \textbf{A$^{H}$} & \textbf{N$^{H}$} &
\textbf{O$^{L}$} & \textbf{C$^{L}$} & \textbf{E$^{L}$} & \textbf{A$^{L}$} & \textbf{N$^{L}$} \\
\midrule

\multirow{4}{*}{\makecell{Motivation}} 
& i) No preferences 
& 33.75 & 20 & 50 & 45 & 32.5 &45 & 30 & 35 & 32.5 & 17.5 & 30 \\

& ii) Mixed-trait preferences 
& 29.5 &27.5	&27.5	&17.5	&42.5	&45	&20	&30	&40	&27.5	&17.5  \\

& iii) Trait-aligned preferences 
& \textbf{88.75}	&85	&92.5	&92.5	&85 &87.5	&92.5	&92.5	&95	&90	&75 \\

& iv) Trait-aligned + 2 noisy preferences 
 &61.75	&52.5	&35	&35	&50	&60	&77.5	&87.5	&87.5	&85	&47.5 \\

\midrule

A. Few-shot 
& Trait-aligned preferences 
& \textbf{88.75}	&85	&92.5	&92.5	&85 &87.5	&92.5	&92.5	&95	&90	&75 \\

\midrule

\multirow{3}{*}{\makecell{B. Few-shot + \\ persona hints}} 
& Preference + pref.\ trait (GT) 
& 89	&97.5&	100	&95&	95	&85&	95	&87.5	&95&	90	&50 \\

& Preference + pref.\ \& choice traits (GT) 
& 32.75	&77.5	&17.5	&72.5	&50&	5	&12.5&	22.5	&42.5	&0	&27.5 \\

& Traits only (pref.\ \& choices) (GT) 
& 8.75&	2.5	&2.5	&15	&5	&0	&0	&5	&10	&0	&47.5 \\

\midrule

C. Reminder 
& Instruction-only (no labels) 
& 74.75	&62.5	&80	&85	&82.5	&82.5	&87.5	&70	&87.5&	62.5	&47.5 \\

\bottomrule
\end{tabular}
\end{small}
\vskip -0.1in
\end{table*}

\begin{table*}[t]
\caption{Accuracy (\%) of Gemini-2.5-pro under different strategies for incorporating persona hints into preference prompting on PACIFIC.}
\label{tab:pacific_results_gemini}
\centering
\begin{small}
\setlength{\tabcolsep}{2pt}
\renewcommand{\arraystretch}{1.05}
\begin{tabular}{p{2cm} p{3cm}| c |ccccc|ccccc}
\toprule
\textbf{Method} & \textbf{Variant} &
\makecell{\textbf{Overall}\\\textbf{Acc. (\%)}} &
\multicolumn{10}{c}{\textbf{Trait-wise Acc. (\%)}} \\
\cmidrule(lr){4-13}
& & &
\textbf{O$^{H}$} & \textbf{C$^{H}$} & \textbf{E$^{H}$} & \textbf{A$^{H}$} & \textbf{N$^{H}$} &
\textbf{O$^{L}$} & \textbf{C$^{L}$} & \textbf{E$^{L}$} & \textbf{A$^{L}$} & \textbf{N$^{L}$} \\
\midrule

\multirow{4}{*}{\makecell{Motivation}} 
& i) No preferences 
&38	&20	&35	&27.5	&50	&47.5	&52.5	&32.5	&40	&22.5	&52.5\\

& ii) Mixed-trait preferences (Total 5 prefs)
&61.5	&55	&72.5	&35	&67.5	&72.5	&65	&52.5	&67.5	&62.5	&65  \\

& iii) Trait-aligned preferences 
&99.25	&100&	100&	100&	100&	100&100	&100	&97.5&	97.5	&97.5 \\

& iv) Trait-aligned + 2 noisy preferences 
&98	&97.5&	100	&100	&100&	95&	100	&97.5	&100&	95&	95 \\

& ii) Mixed-trait preferences (Total 25 prefs)
&59.5	&37.5	&70	&30	&82.5	&65	&85	&35	&77.5	&62.5	&55  \\

\midrule

A. Few-shot 
& Trait-aligned preferences 
&99.25	&100	&100	&100	&100	&100	&100&	100	&97.5	&97.5	&97.5 \\

\midrule

\multirow{3}{*}{\makecell{B. Few-shot + \\ persona hints}} 
& Preference + pref.\ trait (GT) 
&99.25	&100&	100&	100	&100&	95	&100&	100	&97.5	&100	&100 \\

& Preference + pref.\ \& choice traits (GT) 
&99.5	&100	&100	&100	&100	&95	&100	&100	&100	&100	&100 \\

& Traits only (pref.\ \& choices) (GT) 
&97.75	&100	&90	&100	&97.5	&92.5	&100	&100	&100	&97.5	&100 \\

\midrule

C. Reminder 
& Instruction-only (no labels) 
&99.25	&100	&100	&100	&100	&100	&100	&100	&95	&97.5	&100 \\

\bottomrule
\end{tabular}
\end{small}
\vskip -0.1in
\end{table*}

\begin{table*}[t]
\caption{Accuracy (\%) of gpt-4o-mini under different strategies for incorporating persona hints into preference prompting on PACIFIC.}
\label{tab:pacific_results_openai}
\centering
\begin{small}
\setlength{\tabcolsep}{2pt}
\renewcommand{\arraystretch}{1.05}
\begin{tabular}{p{2cm} p{3cm}| c |ccccc|ccccc}
\toprule
\textbf{Method} & \textbf{Variant} &
\makecell{\textbf{Overall}\\\textbf{Acc. (\%)}} &
\multicolumn{10}{c}{\textbf{Trait-wise Acc. (\%)}} \\
\cmidrule(lr){4-13}
& & &
\textbf{O$^{H}$} & \textbf{C$^{H}$} & \textbf{E$^{H}$} & \textbf{A$^{H}$} & \textbf{N$^{H}$} &
\textbf{O$^{L}$} & \textbf{C$^{L}$} & \textbf{E$^{L}$} & \textbf{A$^{L}$} & \textbf{N$^{L}$} \\
\midrule

\multirow{4}{*}{\makecell{Motivation}} 
& i) No preferences 
&50.5	&35	&52.5&	65	&55&	65	&42.5&	75	&55	&12.5&	47.5\\

& ii) Mixed-trait preferences 
&63.5	&52.5&	72.5	&55	&75	&75&	55	&70	&75	&52.5	&52.5  \\

& iii) Trait-aligned preferences 
&97.5	&97.5&	95&	100&	100&	95	&100	&100	&100	&95&	92.5 \\

& iv) Trait-aligned + 2 noisy preferences 
&95	&100&	90	&95&	95&	95	&95	&100	&100&	90	&90 \\

\midrule

A. Few-shot 
& Trait-aligned preferences 
&97.5	&97.5	&95&	100&	100&	95	&100&	100&	100	&95	&92.5 \\

\midrule

\multirow{3}{*}{\makecell{B. Few-shot + \\ persona hints}} 
& Preference + pref.\ trait (GT) 
&99	&100&	100&	100&	100	&97.5	&97.5	&100	&97.5	&97.5	&100 \\

& Preference + pref.\ \& choice traits (GT) 
&99	&100	&100&	100&	100&	100	&97.5	&100&	97.5&	95	&100 \\

& Traits only (pref.\ \& choices) (GT) 
&97.25	&100	&95	&100&	97.5&	100	&85&	100&	97.5	&97.5	&100 \\

\midrule

C. Reminder 
& Instruction-only (no labels) 
&99	&100	&100	&100	&100	&100	&100	&100&	95	&95	&100 \\

\bottomrule
\end{tabular}
\end{small}
\vskip -0.1in
\end{table*}
%%%%%%%%%%%%%%%%%%%%%%%%%%%%%
\section{Hardware Configuration}
\label{sec:hardware configuration}
All experiments were conducted on a high-performance computing cluster equipped with NVIDIA L40S GPUs. The system runs NVIDIA driver version 580.82.07 with CUDA 13.0 support, and PyTorch was compiled with CUDA 12.1. Each GPU provides 46GB of VRAM (46,068 MiB), enabling efficient processing of large language models. The experiments utilized a single GPU per run during model inference.

\section{Methods Description}
\label{app:methods_description}

This section summarizes the prompting templates used in our experiments. Across all methods, we present the user's preference context first and place the target question at the end. We also enforce a strict output format: the model must return a single integer in \{1,2,3,4\}.
\subsection{Few Shot.}
\label{sec:Few-shot}
We provide a set of user preference statements as context and ask the model to predict the user's choice for the target question using \autoref{prompt: preference}.

\begin{floatquote}[h]
\begin{quote}
\noindent
\scriptsize
\begin{center}
\fbox{%
  \parbox{0.95\linewidth}{%
  \vspace{0.6em}
  \texttt{Please analyze the following user preference statements.}\\[0.3em]

    \texttt{<preference>}\\
    \texttt{\{}\\
    \texttt{preference\_statement\_1}\\
    \texttt{preference\_statement\_2}\\
    \texttt{preference\_statement\_3}\\
    \texttt{preference\_statement\_4}\\
    \texttt{preference\_statement\_5}\\
    \texttt{\}}\\
    \texttt{</preference>}\\[0.5em]

    \texttt{Based on these preferences, predict the user’s choice for the following question.}\\[0.3em]

    \texttt{<question>}\\
    \texttt{\{}\\
    \texttt{conversation\_query}\\
    \texttt{Choice 1: ...}\\
    \texttt{Choice 2: ...}\\
    \texttt{Choice 3: ...}\\
    \texttt{Choice 4: ...}\\
    \texttt{\}}\\
    \texttt{</question>}\\[0.5em]

    \texttt{Return a single integer in \{1,2,3,4\}.}
  \vspace{0.6em}
  }%
}
\end{center}
\end{quote}
\caption{Few Shot User Preference prompt}
\label{prompt: preference}
\end{floatquote}

\newpage
\subsection{Noise using unrelated preferences.}
\label{sec:Few-shot-noise}
We provide a set of user preference statements as context and ask the model to predict the user's choice for the target question using \autoref{prompt: Noise using unrelated preferences}.

\begin{floatquote}
\begin{quote}
\noindent
\scriptsize
\begin{center}
\fbox{%
  \parbox{0.95\linewidth}{%
  \vspace{0.6em}
  \texttt{Please analyze the following user preference statements.}\\[0.3em]

    \texttt{<preference>}\\
    \texttt{\{}\\
    \texttt{preference\_statement\_1}\\
    \texttt{preference\_statement\_2}\\
    \texttt{preference\_statement\_3}\\
    \texttt{preference\_statement\_4}\\
    \texttt{preference\_statement\_5}\\
    \texttt{noise\_preference\_statement\_1}\\
    \texttt{noise\_preference\_statement\_2}\\
    \texttt{\}}\\
    \texttt{</preference>}\\[0.5em]

    \texttt{Based on these preferences, predict the user’s choice for the following question.}\\[0.3em]

    \texttt{<question>}\\
    \texttt{\{}\\
    \texttt{conversation\_query}\\
    \texttt{Choice 1: ...}\\
    \texttt{Choice 2: ...}\\
    \texttt{Choice 3: ...}\\
    \texttt{Choice 4: ...}\\
    \texttt{\}}\\
    \texttt{</question>}\\[0.5em]

    \texttt{Return a single integer in \{1,2,3,4\}.}
  \vspace{0.6em}
  }%
}
\end{center}
\end{quote}
\caption{Noise using unrelated preferences}
\label{prompt: Noise using unrelated preferences}
\end{floatquote}

\newpage
\subsection{Few Shot + Persona Trait Hints}
\label{sec:few-shot+persona}
In addition to the preference context, we provide an inferred personality profile (from the preferences) and trait scores for each answer choice using \autoref{prompt:Few Shot + Persona Trait Hints}. The model is instructed to select the choice that best matches both the user's preferences and the provided profile.

\subsection{RAG}

Accuracy for pretrained retriever and fine-tuned retriever per traits are shown in \autoref{tab:rag}.
\begin{table*}[t]
  \caption{Accuracy (\%) for pretrained retriever and fine-tuned retriever}
  \label{tab:rag}
  \centering
  \begin{small}
  \setlength{\tabcolsep}{1.2pt}
  \renewcommand{\arraystretch}{1}
  \begin{tabular}{p{1.2cm} p{2.7cm}| c |ccccc|ccccc}
    \toprule
    \textbf{Method} & \textbf{Variant} &
    \makecell{\textbf{Overall}\\\textbf{Acc. (\%)}} &
    \multicolumn{10}{c}{\textbf{Trait-wise Acc. (\%)}} \\
    \cmidrule(lr){4-13}
    & & &
    \textbf{O$^{H}$} & \textbf{C$^{H}$} & \textbf{E$^{H}$} & \textbf{A$^{H}$} & \textbf{N$^{H}$} &
\textbf{O$^{L}$} & \textbf{C$^{L}$} & \textbf{E$^{L}$} & \textbf{A$^{L}$} & \textbf{N$^{L}$}
 \\
    \midrule

     RAG & pretrained retriever
      & 30.25 & 27.50 & 22.50 & 10.00 & 22.50 & 45.00 & 32.50 & 35.00 & 42.50 & 37.50 & 27.50 \\
          & fine-tuned retriever
      & 43.00 & 32.50 & 30.00 & 32.50 & 37.50 & 52.50 & 40.00 & 55.00 & 62.50 & 52.50 & 35.00 \\
    \bottomrule
  \end{tabular}
  \end{small}
  
  \vskip -0.2in
\end{table*}

\begin{floatquote}
\begin{quote}
\noindent
\scriptsize
\begin{center}
\fbox{%
  \parbox{0.95\linewidth}{%
    \vspace{0.6em}
    \texttt{You are a decision engine that selects the best-matching choice using Big Five (OCEAN) personality traits.}\\[0.3em]

    \texttt{OCEAN definitions:}\\
    \texttt{\{}\\
    \texttt{'O': 'Openness to Experience: creativity, curiosity, willingness to try new ideas',}\\
    ...\\
    \texttt{'N': 'Neuroticism: emotional stability (low) vs.\ anxiety and reactivity (high)'}\\
    \texttt{\}}\\ 

    \texttt{<preference>}\\
    \texttt{\{}\\
    \texttt{preference\_statement\_1}\\
    ...\\
    \texttt{preference\_statement\_5}\\
    \texttt{\}}\\
    \texttt{</preference>}\\ 

    \texttt{User personality profile (inferred from the preferences above):}\\
    \texttt{<user\_profile>}\\
    \texttt{\{}\\
    \texttt{trait: O, level: High}\\
    \texttt{\}}\\
    \texttt{</user\_profile>}\\ 

    \texttt{<question>}\\
    \texttt{\{}\\
    \texttt{conversation\_query}\\
    \texttt{Choice 1: ...}\\
    ...\\
    \texttt{Choice 4: ...}\\
    \texttt{\}}\\
    \texttt{</question>}\\ 

    \texttt{Predicted trait scores for each choice (1--7 per trait):}\\
    \texttt{<choices>}\\
    \texttt{\{}\\
    \texttt{Choice 1: \{O:~, C:~, E:~, A:~, N:~\}}\\
    ...\\
    \texttt{Choice 4: \{O:~, C:~, E:~, A:~, N:~\}}\\
    \texttt{\}}\\
    \texttt{</choices>}\\ 

    \texttt{Task: Choose the single best option (1--4) that best matches the user profile.}\\
    \texttt{Output (STRICT): Return ONLY one integer in \{1,2,3,4\}.}
    \vspace{0.6em}
  }%
}%
\end{center}
\end{quote}
\caption{Few Shot + Persona Trait Hints Prompt}
\label{prompt:Few Shot + Persona Trait Hints}
\end{floatquote}

\newpage

\paragraph{Preferences + Trait Labels of Preferences and Choices.}
We further expose the model’s intermediate signals by providing (i) an OCEAN-based user profile inferred from the preference statements and (ii) predicted OCEAN trait scores for each candidate answer choice shown in \autoref{prompt:prefs-traits}. The model is then instructed to select the option whose trait direction and scores best align with the inferred user profile.
\begin{floatquote}
\begin{quote}
\noindent
\scriptsize
\begin{center}
\fbox{%
  \parbox{0.95\linewidth}{%
    \vspace{0.6em}
    \texttt{You are a decision engine that selects the best-matching choice using Big Five (OCEAN) personality traits.}\\[0.3em]

    \texttt{OCEAN definitions:}\\
    \texttt{\{}\\
    \texttt{'O': 'Openness: creativity, curiosity, willingness to try new ideas',}\\
    ...\\
    \texttt{\}}\\

    \texttt{<preferences>}\\
    \texttt{\{}\\
    \texttt{preference\_statements}\\
    \texttt{\}}\\
    \texttt{</preferences>}\\

    \texttt{User personality profile (inferred from the preferences above):}\\
    \texttt{<user\_profile>}\\
    \texttt{\{}\\
    \texttt{trait\_direction: \{O: High, C: Low, E: High, A: High, N: Low\}}\\
    \texttt{\}}\\
    \texttt{</user\_profile>}\\

    \texttt{<question>}\\
    \texttt{\{}\\
    \texttt{conversation\_query}\\
    \texttt{Choices: ...}\\
    \texttt{\}}\\
    \texttt{</question>}\\ 

    \texttt{Predicted trait scores for each choice (1--7 per trait; include direction when available):}\\
    \texttt{<choices>}\\
    \texttt{\{}\\
    \texttt{option\_1: \{O-high: 7, A-high: 6, N-low: 2\}}\\
    ...\\
    \texttt{option\_4: \{O-low: 2, E-high: 6, N-high: 7\}}\\
    \texttt{\}}\\
    \texttt{</choices>}\\ 

    \texttt{Task: Choose the single best option (1--4) that most closely matches the user profile.}\\
    \texttt{Comparison rules (apply in order):}\\
    \texttt{- Prefer options with the same trait directions (high/low) as the user profile.}\\
    \texttt{- If multiple options match, choose the one with the most matching traits and closest scores.}\\
    \texttt{- If none match clearly, choose the closest overall match given the preferences and choices.}\\[0.4em]

    \texttt{Output (STRICT): Return ONLY one integer in \{1,2,3,4\}. No extra text.}
    \vspace{0em}
  }%
}%
\end{center}
\end{quote}
\caption{Preferences + Trait Labels of Preferences and Choices Prompt.}
\label{prompt:prefs-traits}
\end{floatquote}

\newpage
\paragraph{Preferences' and Choices' Trait Labels Only.}
To isolate the effect of trait signals, we remove the raw preference statements and provide only (i) an inferred OCEAN user profile and (ii) predicted OCEAN trait scores for each answer option using \autoref{prompt:Preferences' and Choices' Trait Labels Only}. The model is instructed to select the option whose trait directions (high/low) and scores best match the user profile.
\begin{floatquote}
\begin{quote}
\noindent
\scriptsize
\begin{center}
\fbox{%
  \parbox{0.95\linewidth}{%
    \vspace{0.6em}
    \texttt{You are a decision engine that selects the best-matching choice using Big Five (OCEAN) personality traits.}\\[0.3em]

    \texttt{OCEAN definitions:}\\
    \texttt{\{}\\
    \texttt{'O': 'Openness: creativity, curiosity, willingness to try new ideas',}\\
    ...\\
    \texttt{'N': 'Neuroticism: emotional stability (low) vs.\ anxiety/reactivity (high)'}\\
    \texttt{\}}\\ 

    \texttt{User personality profile:}\\
    \texttt{<user\_profile>}\\
    \texttt{\{}\\
    \texttt{trait\_direction: \{O: High, C: Low, E: High, A: High, N: Low\}}\\
    \texttt{\}}\\
    \texttt{</user\_profile>}\\ 

    \texttt{You will be given 4 options. Each option includes predicted OCEAN trait scores (1--7).}\\
    \texttt{Select the single best option (1--4) whose trait profile most closely matches the user profile.}\\ 

    \texttt{<choices>}\\
    \texttt{\{}\\
    \texttt{option\_1: \{O-high: 7, A-high: 6, N-low: 2\}}\\
    ...\\
    \texttt{option\_4: \{O-low: 2, E-high: 6, N-high: 7\}}\\
    \texttt{\}}\\
    \texttt{</choices>}\\ 

    \texttt{Comparison rules (apply in order):}\\
    \texttt{- Prefer options with the same trait directions (high/low) as the user profile.}\\
    \texttt{- If multiple options match, choose the one with the most matching traits and closest scores.}\\
    \texttt{- If none match clearly, choose the closest overall match.}\\[0.4em]

    \texttt{Output (STRICT): Return ONLY one integer in \{1,2,3,4\}. No extra text.}
    \vspace{0.6em}
  }%
}%
\end{center}
\end{quote}
\caption{Preferences' and Choices' Trait Labels Only Prompt.}
\label{prompt:Preferences' and Choices' Trait Labels Only}
\end{floatquote}

\newpage
\subsection{Reminder}
\label{sec:reminder}
We use the same prompt as in the few-shot setting, but add a short reminder encouraging the model to consider personality signals implied by the preferences.
\begin{quote}
\scriptsize
\noindent
\begin{center}
\fbox{%
  \parbox{0.95\linewidth}{%
  \vspace{0.6em}
  \texttt{Reminder: Consider the user's personality implied by their preference statements and use it to guide your prediction.}
  \vspace{0.6em}
  }%
}
\end{center}
\end{quote}

Hence, the full prompt is shown in \autoref{prompt:reminder}:
\begin{floatquote}
\begin{quote}
\scriptsize
\noindent
\begin{center}
\fbox{%
  \parbox{0.95\linewidth}{%
  \vspace{0.6em}
  \texttt{Please analyze the following user preference statements.}\\[0.3em]

    \texttt{<preference>}\\
    \texttt{\{}\\
    \texttt{preference\_statement\_1}\\
    ...\\
    \texttt{preference\_statement\_5}\\
    \texttt{\}}\\
    \texttt{</preference>}\\[0.5em]

    \texttt{Based on these preferences, predict the user’s choice for the following question.}\\[0.3em]

    \vspace{0.6em}
  \texttt{Reminder: Consider the user's personality implied by their preference statements and use it to guide your prediction.}
  \vspace{0.6em}

    \texttt{<question>}\\
    \texttt{\{}\\
    \texttt{conversation\_query}\\
    \texttt{Choice 1: ...}\\
    ...\\
    \texttt{Choice 4: ...}\\
    \texttt{\}}\\
    \texttt{</question>}\\[0.5em]

    \texttt{Return a single integer in \{1,2,3,4\}.}
  \vspace{0.6em}
  }%
}
\end{center}
\end{quote}
\caption{Prompt with personality reminder}
\label{prompt:reminder}
\end{floatquote}

\newpage

\section{RAG Fine-tuning}\label{app:rag}
For fine-tuning, we follow the \texttt{biencoder\_local} configuration with the following hyperparameters: batch size $=1$, dev batch size $=16$, Adam $\epsilon=10^{-8}$, Adam betas $(0.9,0.999)$, weight decay $=0.0$, and learning rate $=5\times 10^{-7}$. We fine-tune for 28 epochs until convergence, which takes approximately one hour using the same hardware configuration in Suppl. ~\ref{sec:hardware configuration}. 
% \salma{Don't use Appendix. Use Suppl}

\section{Persona Analysis}
\label{sec:persona analysis}
We evaluate \textbf{32 persona profiles} derived from the Big Five (OCEAN) traits. Since each trait can take one of two levels (\textit{high} or \textit{low}), the full set contains $2^5=32$ distinct personas. Table.~\ref{tab:persona_32_ocean_matrix} lists all personas and their corresponding trait configurations. Using these personas, we compare performance across profiles by measuring \textbf{accuracy on trait-aligned preference prediction} for each persona. The results are summarized below in Figure~\ref{fig:persona_32_ocean}.

\begin{table*}[h]
\centering
\small
\setlength{\tabcolsep}{2pt}
\renewcommand{\arraystretch}{1.05}
\caption{The 32 OCEAN persona profiles (H=High, L=Low). Columns index personas (0--31); rows denote traits.}
\label{tab:persona_32_ocean_matrix}
\resizebox{\textwidth}{!}{%
\begin{tabular}{c|cccccccccccccccccccccccccccccccc}
\toprule
 & \textbf{0} & \textbf{1} & \textbf{2} & \textbf{3} & \textbf{4} & \textbf{5} & \textbf{6} & \textbf{7} &
 \textbf{8} & \textbf{9} & \textbf{10} & \textbf{11} & \textbf{12} & \textbf{13} & \textbf{14} & \textbf{15} &
 \textbf{16} & \textbf{17} & \textbf{18} & \textbf{19} & \textbf{20} & \textbf{21} & \textbf{22} & \textbf{23} &
 \textbf{24} & \textbf{25} & \textbf{26} & \textbf{27} & \textbf{28} & \textbf{29} & \textbf{30} & \textbf{31} \\
\midrule
\textbf{O} & H&H&H&H&H&H&H&H&H&H&H&H&H&H&H&H& L&L&L&L&L&L&L&L&L&L&L&L&L&L&L&L \\
\textbf{C} & H&H&H&H&H&H&H&H& L&L&L&L&L&L&L&L& H&H&H&H&H&H&H&H& L&L&L&L&L&L&L&L \\
\textbf{E} & H&H&H&H& L&L&L&L& H&H&H&H& L&L&L&L& H&H&H&H& L&L&L&L& H&H&H&H& L&L&L&L \\
\textbf{A} & H&H& L&L& H&H& L&L& H&H& L&L& H&H& L&L& H&H& L&L& H&H& L&L& H&H& L&L& H&H& L&L \\
\textbf{N} & H& L&H& L& H& L&H& L& H& L&H& L& H& L&H& L& H& L&H& L& H& L&H& L& H& L&H& L& H& L&H& L \\
\bottomrule
\end{tabular}%
}
\end{table*}

\section{Cross-Validation Results}
\label{app:cross_validation}

To further assess the stability of our benchmark, we conduct a 3-fold cross-validation study on our dataset. As shown in Tab.~\ref{tab:cross_validation_results}, the mean performance remains consistent across folds and aligns with the main results reported in the manuscript. The standard deviation across folds is small for all settings, remaining below 2\%, which confirms the stability of the benchmark.

\begin{table}[t]
\centering
\small
\begin{tabular}{lccccc}
\toprule
\textbf{Preference context} & \textbf{Fold 0} & \textbf{Fold 1} & \textbf{Fold 2} & \textbf{Mean} & \textbf{Std} \\
\midrule
i) Zero-shot & 21.77\% & 25.75\% & 22.73\% & 23.41\% & 1.69\% \\
ii) Mixed Traits & 28.41\% & 24.25\% & 25.00\% & 25.89\% & 1.81\% \\
iii) Aligned Traits & 61.99\% & 66.42\% & 64.02\% & 64.14\% & 1.81\% \\
iv) Contaminated Aligned & 56.46\% & 59.70\% & 59.85\% & 58.67\% & 1.56\% \\
\bottomrule
\end{tabular}
\caption{3-fold cross-validation results on our dataset. The low standard deviation across folds indicates that the benchmark results are stable.}
\label{tab:cross_validation_results}
\end{table}

\section{Memory in Agentic AI for user modeling}
\label{sec: memory discussion}
Prevalent memory architectures, often reduced to RAG-based retrieval, treat user modeling as a naive data storage task. However, this approach could collapse under longitudinal interaction. For scenarios spanning one year of user-LLM interaction, attempting to manage raw history is computationally infeasible and creates reasoning difficulties and noise, making it impossible to effectively retrieve or reason over user preferences amidst the complexity. We argue that effective agentic memory requires shifting from lossless storage to semantic abstraction. It is promising to utilize personality traits as a compression layer; much like a human companion who predicts needs based on ``knowing'' a person rather than recalling every specific event, personality serves as a dense, high-signal proxy that drives user behavior, preference, and decision-making, allowing agents to accurately predict intent without the overhead of exhaustive retrieval.

% Prevalent memory architectures, often reduced to RAG-based retrieval, treat user modeling as a data storage problem. However, our results suggest that this strategy does not scale well in longitudinal interactions.
% When we increased the number of mixed-trait preferences in the LLM context from 5 to 25,  accuracy dropped from 61.5\% to 59.5\% (Suppl.~\ref{app:exp_other_models}, Tab.~\ref{tab:pacific_results_gemini}).
To test this, we scaled the number of mixed-trait preferences provided in the LLM's context from 5 to 25.  Yet, our experiments revealed that expanding the preference context actually degraded LLM accuracy from 61.75\% to 59.5\% (Suppl.~\ref{app:exp_other_models}, Tab.~\ref{tab:pacific_results_gemini}).
This demonstrates that simply feeding an LLM more raw interaction history does not equate to better user modeling; rather, the model becomes overwhelmed by competing signals and noise.
% Therefore, we argue that effective agentic memory requires shifting from lossless storage to semantic abstraction. Much like a human companion who predicts needs based on a synthesized "understanding" of a person rather than recalling every specific event, utilizing personality traits as a dense, high-signal compression layer allows agents to accurately predict intent without the catastrophic noise of exhaustive retrieval.

\section{Limitation \& Future work}

Our study has several limitations that open avenues for future work.

First, we simplify the personality spectrum into a binary High/Low classification. In reality, Big-Five traits are continuous, and individuals sharing the same High or Low direction can still differ in nuanced ways that a binary label cannot capture. In future work, we plan to adopt a finer-grained representation, using the full 7-point Likert scores directly to better reflect the continuous variation in human personality.

Second, our evaluation focuses on multiple-choice question answering. We plan to extend PACIFIC to open-ended generative tasks and compare performance against the QA baseline, assessing whether trait-aligned personalization transfers to free-form generation.

Third, we assume that personality traits are independent of cultural and linguistic context. In practice, the way traits manifest in behavior and preferences may vary across societies, cultures, and languages. We leave a systematic investigation of these cross-cultural and cross-lingual effects to future work.
% \salma{Add more items to your future work. It shouldn't be just more evaluation on models. Maybe what Yu is currently doing.}

\begin{figure}
    \centering
    \includegraphics[width=\linewidth]{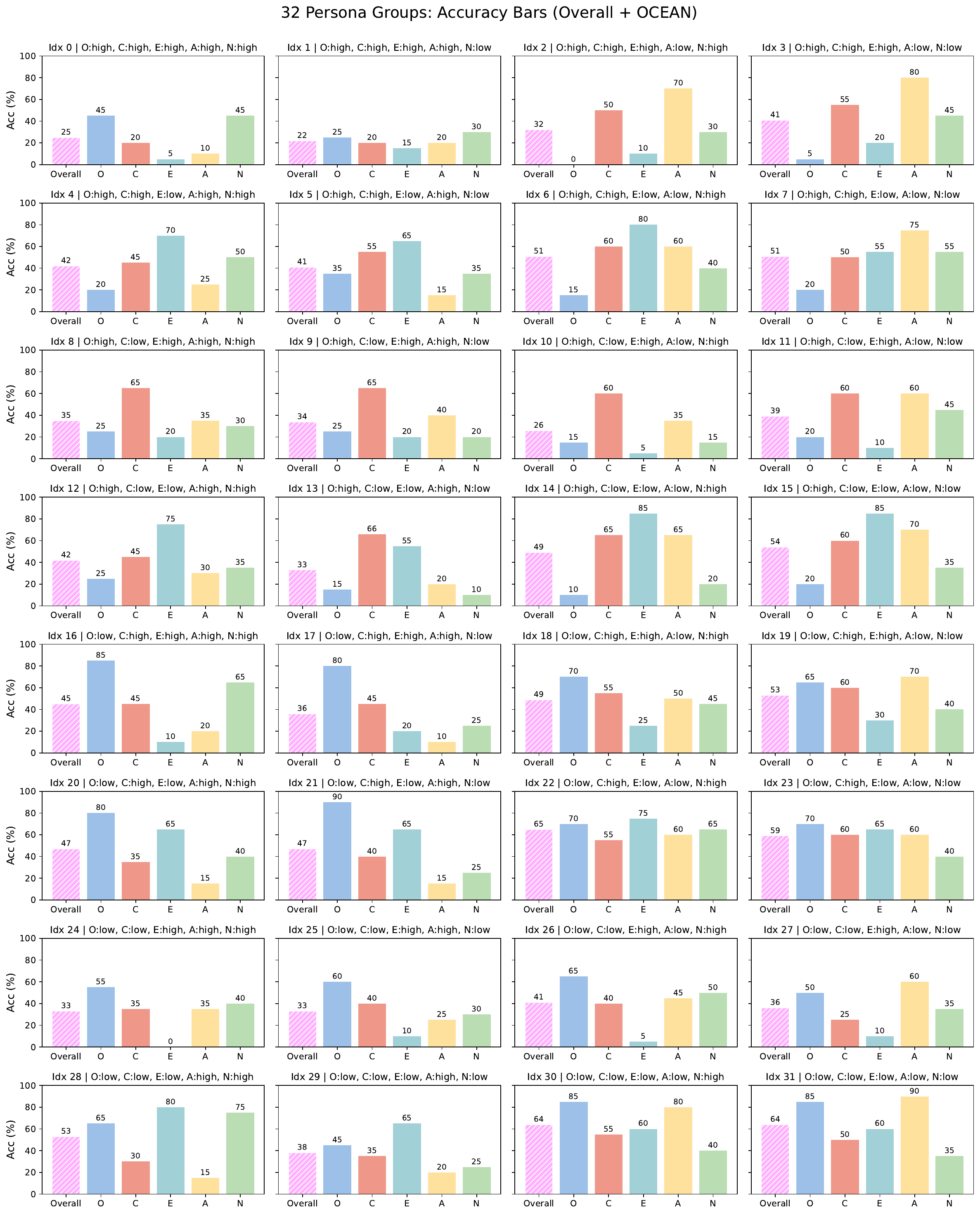}
    \caption{Accuracy (\%) by persona group (32 total). Each panel corresponds to one group (Idx 0–31) and shows the overall accuracy plus per-trait accuracies for the Big Five (O, C, E, A, N). Bar colors encode traits (Overall in magenta with white hatch; O/C/E/A/N in distinct colors), and titles indicate each group’s high/low trait configuration.}
    \label{fig:persona_32_ocean}
\end{figure}

\section*{Impact Statement}
This paper studies how personality trait information (based on the Big Five/OCEAN framework) can improve an LLM’s ability to follow user preferences in multiple-choice question answering, and proposes retrieval-based methods for selecting trait-aligned preference statements when annotations are unavailable. The intended positive impact is to advance research on controllable and user-aligned generation, which may help build assistants that are more consistent with a user’s stated goals and reduce frustrating or irrelevant responses.

We view this work primarily as a measurement and method-development contribution, and we encourage responsible use. We do not foresee specific ethical concerns or negative societal consequences arising from this work beyond those commonly associated with improving the capability and controllability of machine learning models. To support transparency and reproducibility, we plan to release our dataset and code, and we follow standard research practices and a code of conduct for responsible use. Overall, we expect this work to contribute to more personalized and helpful LLM systems in the future, improving user experience and better assisting people in everyday tasks.

%You can have as much text here as you want. The main body must be at most $8$ pages long. For the final version, one more page can be added. If you want, you can use an appendix like this one.

%The $\mathtt{\backslash onecolumn}$ command above can be kept in place if you prefer a one-column appendix, or can be removed if you prefer a two-column appendix.  Apart from this possible change, the style (font size, spacing, margins, page numbering, etc.) should be kept the same as the main body.
%%%%%%%%%%%%%%%%%%%%%%%%%%%%%%%%%%%%%%%%%%%%%%%%%%%%%%%%%%%%%%%%%%%%%%%%%%%%%%%
%%%%%%%%%%%%%%%%%%%%%%%%%%%%%%%%%%%%%%%%%%%%%%%%%%%%%%%%%%%%%%%%%%%%%%%%%%%%%%%

\end{document}